\RequirePackage{fix-cm}
\documentclass[smallcondensed]{svjour3}     
\smartqed  
\usepackage{graphicx}
\usepackage{graphics}
\usepackage{amsfonts}
\usepackage{amstext}
\usepackage{amsmath}
\usepackage{amssymb}
\usepackage{enumerate}
\usepackage{mathtools}
\usepackage{newlfont}
\usepackage{tabularx}
\usepackage{lscape}
\usepackage{multirow}
\usepackage{epsfig}
\usepackage{bm}
\usepackage{algorithm}
\usepackage{algorithmic}
\usepackage{url}
\usepackage{xcolor}
\usepackage{multirow}
\usepackage{paralist}
\usepackage[square,sort,comma,numbers]{natbib}

\usepackage[misc]{ifsym}
\begin{document}
	
	\title{{Multi-Kernel Fusion for RBF Neural Networks}}
	
	
	\author{Syed~Muhammad~Atif\protect\footnotemark \and Shujaat~Khan\footnotemark[\value{footnote}] \and Imran~Naseem \and 			
		Roberto~Togneri \and Mohammed~Bennamoun \footnotetext{Both authors contributed equally.}}
	
	
\institute{Syed Muhammad Atif \at 
	Graduate School of Science and Engineering, Karachi Institute of Economics and Technology, Korangi Creek, Karachi 75190, Pakistan.\\
	\email{s.m.atif@pafkiet.edu.pk} 
	\and	
	Shujaat Khan \at
	Department of Bio and Brain Engineering,
	Korea Advanced Institute of Science and Technology (KAIST), Daejeon 34141,
	Republic of Korea. \\
	\email{shujaat@kaist.ac.kr}           
	\and
	Imran Naseem\Letter \at 
	College of Engineering, Karachi Institute of Economics and Technology, Korangi Creek, Karachi 75190, Pakistan.  \\School of Electrical, Electronic and Computer Engineering, The University of Western Australia, 35 Stirling Highway, Crawley, Western Australia 6009, Australia.\\
	\emph{\email{imran.naseem@uwa.edu.au}}
	\and
	Roberto Togneri \at School of Electrical, Electronic and Computer Engineering, The University of Western Australia, 35 Stirling Highway, Crawley, Western Australia 6009, Australia.\\
	\email{roberto.togneri@uwa.edu.au}
	\and
	Mohammed Bennamoun\at
	School of Computer Science and Software Engineering, The University of Western Australia, 35 Stirling Highway, Crawley, Western Australia 6009, Australia.\\  \email{mohammed.bennamoun@uwa.edu.au}
}
	
	\date{Received: date / Accepted: date}

	\maketitle

\begin{abstract}    
	A simple yet effective architectural design of radial basis function neural networks (RBFNN) makes them amongst the most popular conventional neural networks. The current generation of radial basis function neural network is equipped with multiple kernels which provide significant performance benefits compared to the previous generation using only a single kernel. In existing multi-kernel RBF algorithms, multi-kernel is formed by the convex combination of the base/primary kernels. In this paper, we propose a novel multi-kernel RBFNN in which every base kernel has its own (local) weight. This novel flexibility in the network provides better performance such as faster convergence rate, better local minima and resilience against stucking in poor local minima. These performance gains are achieved at a competitive computational complexity compared to the contemporary  multi-kernel RBF algorithms. The proposed algorithm is thoroughly analysed for performance gain using mathematical and graphical illustrations and also evaluated on three different types of problems namely: 
		\begin{inparaenum}[(i)]
		\item pattern classification, 
		\item system identification and 
		\item function approximation.
		\end{inparaenum}
	Empirical results clearly show the superiority of the proposed algorithm compared to the existing state-of-the-art multi-kernel approaches.
\end{abstract}

\section{Introduction}\label{introduction}

Machine learning (ML) is an established field with a wide range of applications including control engineering \cite{de_almeida_rego_deterministic_2014,khan2018fractional,meng_nonlinear_2018,ibrahim2020machine}, medical imaging \cite{khan2020adaptive,yoon2018efficient,pratiwi_mammograms_2015}, bioinformatics \cite{khan2018rafp,naseem2017ecmsrc,usman2020afp}, and design of forecasting systems \cite{gan_hybrid_2012,zhu_traffic_2014,sadiq2018chaotic,khan2018novel}, etc. It has been successfully used for other innovative applications as well such as in the design of cognitive communication systems \cite{bu2020adversarial,peng2018modulation} and powerful generative models for number of multimedia application \cite{lee2019collagan,goodfellow2016nips} . In ML, neural networks are considered to be an important category of tools being frequently used. Therefore number of neural network architectures for example spiking neural neural network (SPNN),  multiple layer perceptron (MLP), convolutional neural networks (CNN) and radial basis function neural network (RBFNN) has been proposed. 

Due to its compact design and good noise tolerance RBFNN is extensively used in various applications where computational complexity, and data availability is a constrain \cite{aljarah_training_2018}. Several advances have been proposed to improve its performance. For instance, to improve the parameter learning a variant of gradient decent has been proposed \cite{khan2018fractional}, instead of gradient descent algorithms some researchers have used meta-heuristic algorithms to update kernel weights and other network parameters \cite{aljarah_training_2018,alexandridis_cooperative_2016,yang_nature-inspired_2010,simon_biogeography-based_2008}. Aljarah \textit{et al.} in \cite{aljarah_training_2018}, used bio-geography-based optimization algorithm (BBO) \cite{simon_biogeography-based_2008}. Alexandridis \textit{et al.} studied the effectiveness of particle swarm algorithm (PSO) for updating weights of the RBFNN \cite{alexandridis_cooperative_2016}. 

Recently researchers have successfully blended RBFNN with other established techniques as well. For example \cite{wen_robust_2019,yang_fast_2018,liu_c-rbfnn:_2018},
Yang et. al in \cite{yang_fast_2018} proposed an efficient method for the selection of the centers using the conventional K-means clustering. However, unnecessary points around cluster centers were removed during global K-means clustering using population density method. This slight tweak in the selection procedure of the center, resulted in faster convergence and more robustness. 
In \cite{wen_robust_2019}, Wena et. al used Takagi–Sugeno (TS) fuzzy model with the RBF neural network. The proposed designed is particularly useful in environments with data loss, data distortion or signal saturation. It uses K-means clustering for both selecting fuzzy rules and the centers of the RBFNN. Moreover, weighted activation degree (WAD) is used to determine the firing strength of fuzzy node.  
Liu et. al \cite{liu_c-rbfnn:_2018} proposed C-RBFNN (Cloud RBFNN) which uses the cloud theory in fuzzy mathematics to optimize the activation functions. This modification allows RBFNN to effectively express the fuzziness and randomness of the user data such as social media data.

Some hybrid training options have also been recently explored. For instance in \cite{chen_combining_2019}, Yao and Kuo proposed to combine self-organizing map (SOM) based RBF with evolutionary algorithms such as partical swarm optimization (PSO) and genetic algorithm (GA). This hybrid approach for RBF outperformed conventional non-hybrid approaches. Another emerging variant of RBFNN called spatio-temporal RBFNN, uses the concept of time-space orthogonality to separately model the dynamics and nonlinear complexities \cite{khan_spatio-temporal_2018,sadiq2018chaotic}. Additionally, an adaptive Nelder Mead Simplex \cite{gao_implementing_2012}, based training method that simultaneously updates weights and kernel width is proposed in \cite{hassan_kernel_2018}. 

\subsection{Motivation and contribution of this research}
RBFNN typically uses a single type of kernel lacking better generalization. This is because practical learning problems often involve multiple, heterogeneous data sources. Hence, the choice of kernel is heavily dependent on the problem at hand \cite{fu_sparse_2010,aftab_novel_2014}.  For example, wavelet kernel, due to its excellent local properties both in time and frequency domains, performs better for some signal approximation and pattern classification problems, however due to lack of prior knowledge choosing the best kernel for the given learning problem is a challenging task. An alternative approach is to use multiple kernels to incorporate design flexibility and generalization \cite{fu_sparse_2010,bucak_multiple_2014,varma_more_2009}. This approach has been successfully employed with other kernel-based methods for instance in support vector machine (SVM) \cite{tuia_learning_2010,vetrivel_disaster_2018}. The most widely used approach to combine multiple kernels of different characteristics is convex combination i.e. all participating kernels are combine linearly such that their coefficients are non-negative and sum to unity \cite{tuia_learning_2010,vetrivel_disaster_2018,muhammad_weighted_2017}. Recently, some researchers have made successful attempts to combine multiple kernels in a nonlinear fashion e.g. Gu, Yanfeng, et al. in \cite{gu_nonlinear_2016} showed the effectiveness of combining multiple kernels using Hadamard product.

{In the context of RBFNN, multi kernel approach is still an under-explored research area. Fu et al. \cite{fu_sparse_2010} were the first to introduce the multi kernel RBF-NN. They combined the Gaussian kernel and the wavelet kernel using convex combination and adaptively tuned the kernel coefficients using orthogonal least squares (OLS) algorithm. Later, Aftab \textit{et al.} in \cite{aftab_novel_2014} and Khan \textit{et al.} in \cite{khan2017novel} explored the area of multi-kernel RBFNN and designed an adaptive multi-kernel RBFNN. Motivated from these works, we propose a novel muti-kernel RBFNN architecture as a Coordinating RBF Neural Network (Co-RBFNN).} 

{Conventional multi-kernel RBF architectures, use the concept of linear combination of various primary kernels (Gaussian, cosine, wavelet etc) with either fixed or adaptive weights, incorporating single degree of freedom \cite{fu_sparse_2010,aftab_novel_2014,khan2017novel}.  In particular, the conservative choice of the mixing parameters turns out to be the limitation of these conventional approaches. In contrast, the proposed kernel fusion method uses matrix-based mixing weights allowing each participating kernel to learn independently, thereby yielding better performance in most cases. This learning approach of independent mixing weights, make our method novel and unique compared to other contemporary approaches.}  
{
The main contributions of our research are as follows: 
\begin{enumerate}
	\item A multi-kernel RBFNN architecture is proposed that combines each multi-kernel in the network with its own set of kernel parameters (local weights).
	\item Graphical explanation of the algorithm is given to conceptually justify the origin of improved performance.
	\item A comprehensive mathematical analysis is performed to identify the convergence bound.
	\item The proposed architecture is evaluated for three problems of estimation namely non-linear system identification, pattern classification, and function approximation and extensive comparative analysis is performed with the contemporary approaches.
\end{enumerate}
}
The organization of the paper is as follows.  In section \ref{sec:overview}, a brief overview of existing multi-kernel RBFNNs is proposed followed by the proposed Co-RBFNN in section \ref{sec:proposed}. Experimental evaluation and comparative results are discussed in section \ref{sec:results}. Finally, the paper is concluded in Section \ref{sec:conclusion}. 

\section{Multi-Kernel Radial Basis Function Neural Networks}
\label{sec:overview}
\subsection{Overview of the architecture of the RBF neural network}
\begin{figure}[ht!]
	\begin{center}
		\centering 
		\includegraphics*[scale=0.3]{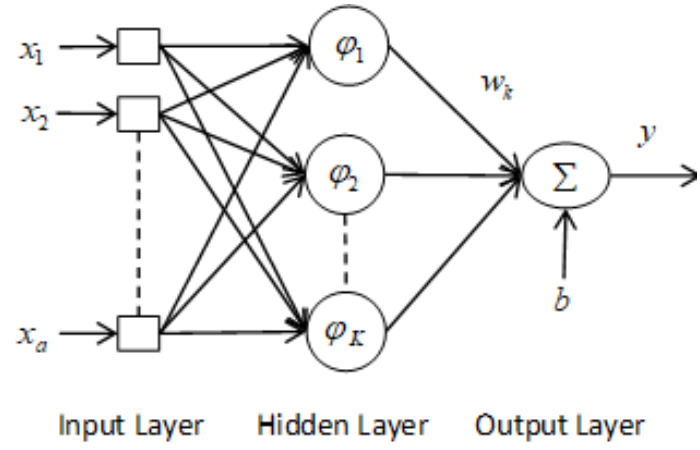}
	\end{center}
	\caption{Architecture of the RBF neural network.}
	\label{rbfarch}
\end{figure}

RBFNN is a simple feed forward neural network that consists of only three layers i.e., an input layer, a nonlinear hidden layer and a linear output layer. Fig.~\ref{rbfarch} depicts the architecture of an RBFNN. Let $\mathbf{X} \in \mathbb{R}^{a\times S}$ representing an input dataset consist of $S$ samples, and $\mathbf{x}_s \in \mathbb{R}^{a\times1}$ be the input vector representing a sample by $a$ number of attributes, then the overall mapping of the RBF network, $f:\mathbb{R}^{a\times1}\rightarrow\mathbb{R}^{1\times1}$, is given as:
\begin{eqnarray}\label{eq:simple_rbf}
y_s=\sum_{k=1}^{K}w_k\phi_k(\mathbf{x}_s,\mathbf{m}_k)+b,
\end{eqnarray} 
where for all $k$, $\mathbf{m}_k \subset \mathbf{M} \in \mathbb{R}^{a\times K}$, $K$ is the number of neurons in the hidden layer of the network, $\mathbf{M} \in \mathbb{R}^{a\times K}$ comprises of $K$ number of $\mathbf{m}_k \in \mathbb{R}^{a\times1}$ vectors, each representing a center point of the kernel of $k^{th}$ hidden neuron, $\mathbf{w}_k$ is the synaptic weight connecting the $k^{th}$ hidden neuron to the output neuron, $b$ is the bias term of the output neuron and $\phi_k$ is the radial basis function of the $k^{th}$ hidden neuron.  Without the loss of generality and for the sake of simplicity a single output neuron is considered.  Conventional RBF networks employ a number of kernels such as multiquadrics, inverse multiquadrics and Gaussian \cite{haykin_neural_1999}.
  
\subsection{Overview of the contemporary multi-kernel approaches}
Gaussian kernel is considered to be the most commonly used kernel: 
\begin{eqnarray}\label{eq:gaussian_kernel}
\phi_g(\mathbf{x},\mathbf{m})=\exp\left(\frac{-\left\|\mathbf{x}-\mathbf{m}\right\|^2}{\sigma^{2}}\right),
\end{eqnarray}   
where $\sigma$ is the kernel-width of the Gaussian kernel.

Recently, it has been argued that the cosine kernel offers complimentary information compared to the Gaussian kernel \cite{aftab_novel_2014}. It is defined as:  
\begin{eqnarray}\label{eq:cosine_kernel}
\phi_{c}(\mathbf{x},\mathbf{m})=\frac{\mathbf{x}.\mathbf{m}}{\left\|\mathbf{x}\right\|\left\|\mathbf{m}\right\| + \epsilon},
\end{eqnarray} 
where, $\|| \cdot \||$ is the L2 norm or Euclidean distance and $\epsilon > 0$ is a small constant added to avoid the indeterminant form of Eq\eqref{eq:cosine_kernel}.

In recent studies \cite{gu_nonlinear_2016,bucak_multiple_2014,tuia_learning_2010,varma_more_2009}, it is suggested that combining multiple kernels is more efficient than using the kernels individually. Accordingly, a novel multi-kernel has been proposed combining cosine and Gaussian kernels \cite{aftab_novel_2014}:
\begin{eqnarray}\label{eq:fixed_multi-kernel}
\phi_k(\mathbf{x},\mathbf{m}_k)=\alpha_{g}\phi_{g}(\mathbf{x},\mathbf{m}_k)+\alpha_{m}\phi_{c}(\mathbf{x},\mathbf{m}_k),
\end{eqnarray}
where $\phi_{g}(\mathbf{x},\mathbf{m}_k)$ and $\phi_{c}(\mathbf{x},\mathbf{m}_k)$ are output of Gaussian and cosine kernels for $k^{th}$ hidden neuron respectively and, $\alpha_{g}$ and $\alpha_{c}$ are their corresponding kernel weights. Further, there are two constraints on $\alpha_{g}$ and $\alpha_{c}$, i.e., $0 \leq \alpha_{g},\alpha_{c} \leq1$ and $\alpha_{g}+\alpha_{c}=1$. The common set of kernel weights i.e., $\{\alpha_{g}, \alpha_{c}\}$ for all multi-kernels and the above two constraints ensures that the participating kernels will form a convex combination. 
  
The new multi-kernel in \eqref{eq:fixed_multi-kernel} has shown some good results compared to the conventional Gaussian kernel \cite{aftab_novel_2014}. In this method, the fusion of the two kernels is manual and the their weights $\alpha_{g}$ and $\alpha_{c}$ are adjusted in a hit-and-trial manner. Without any prior information, a common practice is to assign equal weights to the two kernels i.e. $\alpha_{g}=\alpha_{c}=0.5$. To resolve this issue, in \cite{khan2017novel}, an adaptive framework is proposed for automatic fusion of kernels. This approach tunes the kernel weights at every iteration $n$ to minimize error \cite{khan2017novel}:
\begin{eqnarray}\label{eq:adaptive_multi-kernel}
\phi_k(\mathbf{x},\mathbf{m}_k)=\alpha_{g}(n)\phi_{g}(\mathbf{x},\mathbf{m}_k)+\alpha_{c}(n)\phi_{c}(\mathbf{x},\mathbf{m}_k).
\end{eqnarray}

In \cite{khan2017novel}, both the synaptic weights of hidden neuron and kernel weights are updated using the conventional gradient descent algorithm. This method has shown improvement over the fixed multi-kernel methods\cite{aftab_novel_2014}. 

\section{The proposed Coordinating RBFNN (Co-RBFNN)}
\label{sec:proposed}
Motivated by \cite{khan2017novel}, we argue that this adaptive scheme can be further improved by introducing a separate set of kernel weights for each participating kernel. Therefore, the $k^{th}$ kernel of the given RBFNN that consists of two participating kernels will take the form:
\begin{eqnarray}\label{eq:robust_multi-kernel}
		\phi_k(\mathbf{x},\mathbf{m}_k)&=\alpha_{g_k}(n)\phi_{g}(\mathbf{x},\mathbf{m}_k)+\alpha_{c_k}(n)\phi_{c}(\mathbf{x},\mathbf{m}_k),
\end{eqnarray}
where $\phi_{g_k}(\mathbf{x},\mathbf{m}_k)$ and $\phi_{c}(\mathbf{x},\mathbf{m}_k)$ are the Gaussian and cosine contributors of the $k^{th}$ multi-kernel with the corresponding weights $\alpha_{g_k}(n)$ and $\alpha_{c_k}(n)$ respectively. 
Eq \eqref{eq:robust_multi-kernel} can be rewritten as:  
\begin{eqnarray}\label{eq:2nd-robust_multi-kernel}
\phi_k(\mathbf{x},\mathbf{m}_k)=\sum_{l}^{L}\alpha_{l_k}(n)\phi_{l_k}(\mathbf{x},\mathbf{m}_k),
\end{eqnarray}
where, $l \in L$ and $L= \{g,c\}$ is the set of participating primary kernels in the $k^{th}$ multi-kernel. So, $\phi_{l_k}$ is the $l^{th}$ participating primary kernel of the $k^{th}$ kernel and $\alpha_{l_k}$ is its mixing weight.

Eq\eqref{eq:2nd-robust_multi-kernel} can be easily extended for more than two kernels. However, we restrict ourselves to only two kernels for the sake of simplicity. 
The overall mapping at the $n^{th}$ iteration can be written as:
\begin{eqnarray}\label{eq:robust-rbf-nn}
y(n)=\sum_{k=1}^{K}w_{k}(n)\Bigg(\sum_{l \in \{g,c\}}\alpha_{l_k}(n)\phi_{l_k}(\mathbf{x}(n),\mathbf{m}_k)\Bigg)+b(n),
\end{eqnarray} 
where $K$ is the number of centers (multi-kernel) of the network, $\mathbf{m}_k \in \mathbb{R}^{a\times1}$ is the center of the $k^{th}$ multi-kernel, $\mathbf{w}_k$ is the synaptic weight connecting the $k^{th}$ hidden neuron to the output neuron, $b$ is the bias term of the output neuron, $\phi_{l_k}$ is the $l^{th}$ participating kernel of $k^{th}$ multi-kernel and $\alpha_{l_k}$ is the corresponding kernel weight.

Eq\eqref{eq:robust-rbf-nn} can be written as:
\begin{eqnarray}\label{eq:2nd-robust-rbf-nn}
\begin{split}
y(n)&=\sum_{k,l}\Bigg(w_{k}(n)\alpha_{l_k}(n)\Bigg)\phi_{l_k}(\mathbf{x}(n),\mathbf{m}_k)+b(n)\\
	&=\sum_{k,l}w_{k,l}(n)\phi_{l_k}(\mathbf{x}(n),\mathbf{m}_k)+b(n),
\end{split}
\end{eqnarray}
where, $k=1,2,..., K$, $l \in \{g,c\}$ and $w_{k,l}(n) = w_{k}(n)\alpha_{l_k}(n)$ is the substitute form of the weight of $l^{th}$ participating kernel in the $k^{th}$ multi-kernel. $\mathbf{x}(n)$ is a sample obtained from $\mathbf{X}$ at $n^{th}$ iteration.

It is evident from Eq\eqref{eq:2nd-robust-rbf-nn} that there is no explicit need to maintain kernel weight of each participating kernel of a given multi-kernel. Instead, each participating kernel $\phi_{l_k}$ has its own corresponding weight $w_{k,l}(n)$. In other words, our proposed multi-kernel RBFNN architecture, consisting of $K$ hidden neurons and $L$ participating kernels (in our case $L=2$), may be unfolded into a simple RBFNN architecture consisting of $K \times L$ centers (hidden neurons), such that there are $L$ sets of $K$ hidden neurons and each of that set employs one of the $L$ different kernels.   

In matrix form, Eq\eqref{eq:2nd-robust-rbf-nn}  can be written as:
\begin{eqnarray}\label{eq:matrix-robust-rbf-nn}
	y(n)=\bm{\phi}^{\intercal}(n)\bm{w}(n),
\end{eqnarray}
where, $\bm{w}(n) = [b, w_{g_1}(n), w_{g_2}(n), \cdots, w_{g_K}(n), w_{c_1}(n), w_{c_2}(n), \cdots, w_{c_K}(n)]^{\intercal}$ and $\bm{\phi}(n) = [1, \phi_{g_1}(\mathbf{x}(n),\mathbf{m}_k), \cdots, \phi_{g_K}(\mathbf{x}(n),\mathbf{m}_k), \phi_{c_1}(\mathbf{x}(n),\mathbf{m}_k), \cdots, \phi_{c_K}(\mathbf{x}(n),\mathbf{m}_k)]^{\intercal}$ are weights and kernel vectors respectively and $[\cdot]^{\intercal}$ is the vector transpose operation. 

\subsection{Weight and bias update rules}
The update rule of the synaptic weight $w_{k,l}(n)$ at $(n+1)^{th}$ iteration can be given as: 
\begin{eqnarray}\label{weight_update_rule}
w_{k,l}(n+1)=w_{k,l}(n)+\Delta w_{k,l}(n),
\end{eqnarray}
\begin{eqnarray}\label{delta_w}
\Delta w_{k,l}(n)=-\eta\frac{\partial \mathcal{\ell}}{\partial w_{k,l}(n)},
\end{eqnarray}
where, $\eta$ is the learning rate, and $\ell$ is the mean-square-error ($L2$) loss function defined as:
\begin{eqnarray}\label{l2_cost_function}
\mathcal{\ell}\left(\bm{w},b\right)=\frac{1}{N}\sum_{n=1}^{N}(d(n)-y(n))^{2}.
\end{eqnarray} 

The above loss function can be minimized by solving for the instantaneous error, considering instantaneous error function $\mathcal{E}(n)$ i.e.,:

\begin{eqnarray}\label{cost_function}
\mathcal{E}(n)=\mathcal{E}\left(\bm{w}(n),b(n)\right)=\frac{1}{2}(d(n)-y(n))^{2},
\end{eqnarray} 
where $d(n)$ is the desired output, $y(n)$ is the actual output at the $n^{th}$ iteration and $e(n)$ the instantaneous error.\\
Using the chain rule of differentiation for the cost function in Eq\eqref{cost_function} yields:
	\begin{eqnarray}\label{partial_w}
		\frac{\partial \mathcal{E}(n)}{\partial w_{k,l}(n)}=\frac{\partial \mathcal{E}(n)}{\partial e(n)}\frac{\partial e(n)}{\partial y(n)}\frac{\partial y(n)}{\partial w_{k,l}(n)},
	\end{eqnarray}
which upon simplification of the partial derivatives in Eq\eqref{partial_w} results in:
	\begin{eqnarray}\label{simplified_partial_w}
		\frac{\partial \mathcal{E}(n)}{\partial w_{k,l}(n)}=-e(n)\phi_{l_k}(\mathbf{x}(n),\mathbf{m}_k).
	\end{eqnarray}

Using Eq\eqref{delta_w} and Eq\eqref{simplified_partial_w}, the update rule in  Eq\eqref{weight_update_rule} will becomes:
\begin{eqnarray}\label{eq:simplified_weight_update_rule}
w_{k,l}(n+1)=w_{k,l}(n)+\eta e(n)\phi_{l_k}(\mathbf{x}(n),\mathbf{m}_k),
\end{eqnarray}
similarly, the update rule for bias $b(n)$ can be shown to have the form:
\begin{eqnarray}\label{eq:simplified_bias_update_rule}
b(n+1)=b(n)+\eta e(n).
\end{eqnarray}

{\subsection{Training algorithm:}}\label{sec:training_algo}

{For the training of the proposed network, the steps of the algorithm outlined in Table~\ref{algo:training_co-rbf} are followed. Define the inputs, $X \in \mathbb{R}^{a \times S}$, $M \in \mathbb{R}^{a \times K}$ (where the columns are the centers of the $K$ multi-kernels) the initial weight matrix $W_{init} \in \mathbb{R}^{K\times L}$, initial value of bias $b$, the learning rate $\eta > 0$ and T number of epochs for training. The algorithm yields a weight matrix $W \in \mathbb{R}^{K \times L}$ as output. Conventional stochastic gradient descent is used to update the weight matrix $W \in \mathbb{R}^{K \times L}$ independently using each of the $S$ training samples in each of the $T$ epochs.}

	\begin{table}[!h]
		\caption{Algorithmic depiction of the proposed Co-RBFNN.
		\label{algo:training_co-rbf}}
		\begin{center}
			\begin{tabular}{l}
			\hline
			\textbf{Require:} An $a$-by-$S$ training data matrix $X$ consist of $S$ samples of $a$ dimension\\
			(features/attributes), an $1$-by-$S$ training desired response matrix $d$ of corresponding\\
			$S$ samples, $K$-by-$L$ kernel functions $\Phi$, $a$-by-$K$ matrix $M$ for $K$ multi-kernel centers\\
			(means), an $K$ by $L$ initial weight matrix $W_{init}$, initial bias $b_{init}$, $\eta$ the learning rate\\
			for the weights and bias, and $T$ number of training epochs.\\
			\textbf{Ensure:} an $K$ by $L$ final weight matrix $W$\\
			\textbf{Initialize:} $W = W^{(prev)} = W_{init}$; $b = b^{(prev)} = b_{init}$; $t = 1$;\\
		
    		\textbf{repeat}\\
    		    $\qquad s = 1$;\\			
    		    \qquad \textbf{repeat}\\
    		        $\qquad \qquad k = 1$; $y_{s} = b$;\\
    		        \qquad \qquad \textbf{repeat}\\
    		            $\qquad \qquad \qquad l = 1$;\\
    		            \qquad \qquad \qquad \textbf{repeat}\\
    		                $\qquad \qquad \qquad \qquad y_{s} = y_{s} + \bm{w}_{k,l}\phi_{l_k}(\mathbf{x}_s,\mathbf{m}_k)$;\\
    		                $\qquad \qquad \qquad \qquad l = l + 1$;\\
    		            \qquad \qquad \qquad \textbf{until} $l \le L$;\\
    		            $\qquad \qquad \qquad k = k + 1$;\\
    		        \qquad \qquad \textbf{until} $k \le K$;\\
    		        $\qquad \qquad e_{s} = d_{s} - y_{s}$;\\
    		        $\qquad \qquad W^{(prev)} = W$; $b^{(prev)} = b$;\\
    		        $\qquad \qquad k = 1$;\\
    		        \qquad \qquad \textbf{repeat}\\
    		            $\qquad \qquad \quad l = 1$;\\
    		            \qquad \qquad \quad     \textbf{repeat}\\
    		                $\qquad \qquad \qquad \qquad \bm{w}_{k,l} = \bm{w}_{k,l}^{(prev)} + \eta e_{s} \phi_{l_k}(\mathbf{x}_s,\mathbf{m}_k)$;\\
    		                $\qquad \qquad \qquad \qquad l = l + 1$;\\
    		            \qquad \qquad \quad \textbf{until} $l \le L$;\\
    		            $\qquad \qquad \qquad k = k + 1$;\\
    		        \qquad \qquad \textbf{until} $k \le K$;\\
    		        $\qquad \qquad b = b^{(prev)} + \eta e_{s}$;\\
    		        $\qquad \qquad s = s + 1$;\\
    		    \qquad \textbf{until} $s \le S$\\ 
    		    $\qquad t = t + 1$;\\
    		\textbf{until} $t \le T$\\
			\hline
			\end{tabular} 	
		\end{center}
	\end{table}

\subsection{Illustrative explanation of the proposed method}\label{sec:explanation_co-rbf}
\begin{figure}[ht!] 
		\centering
		\includegraphics[width=0.7\textwidth]{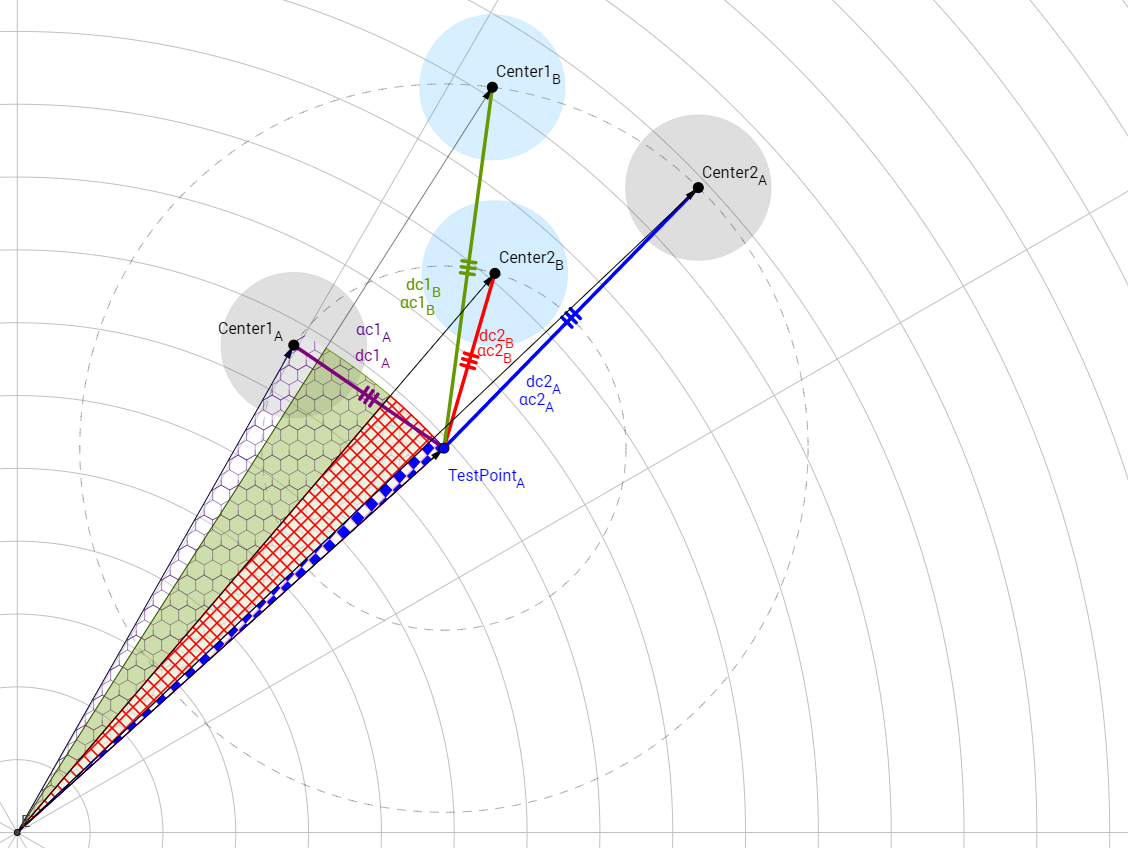}
		\caption{Illustrative explanation of the proposed RBF algorithm. \label{fig:RBF_GE}} 
	\end{figure}

	{In this subsection, we consider an illustrative example depicted in {Fig.} \ref{fig:RBF_GE}. The task is to classify a test point. It is illustratively proved that a primary kernel (which is a Gaussian or a cosine kernel in this example) fails to effectively discriminate the given test point. In contrast, our proposed solution effectively maps the given test point to its true class. This illustration therefore serve to demonstrate the superiority of the proposed method. For the purpose of this illustrative case-study, no assumptions were made except the choice of a highly challenging test point to prove the efficacy of the proposed algorithm for difficult cases.}

	As depicted in Fig.~\ref{fig:RBF_GE}, we consider a challenging binary classification problem, in which the only tunable parameters are the kernel mixing weights. We have four center points obtained using a clustering method such as K-mean clustering (or any other method) representing two classes namely $ClassA$ and $ClassB$.  As shown in Fig.~\ref{fig:RBF_GE}, $Center1_{A}$ and $Center2_{A}$ are the representative points of $ClassA$ and $Center1_{B}$ and $Center2_{B}$ are the representative points of $ClassB$ respectively. Let's consider a test sample $TestPoint_{A}$ such that $dc1_{A}$, $dc2_{A}$ are Euclidean distances from $TestPoint_{A}$ to centers $Center1_{A}$ and $Center2_{A}$ respectively whereas $dc1_{B}$, $dc2_{B}$ are Euclidean distances of test sample $TestPoint_{A}$ from centers $Center1_{A}$ and $Center2_{A}$ respectively. Similarly, $ac1_{A}$, $ac2_{A}$ are angles of test sample $TestPoint_{A}$ with centers $Center1_{A}$ and $Center2_{A}$ respectively whereas $ac1_{B}$, $ac2_{B}$ are angles of test sample $TestPoint_{A}$ with centers $Center1_{B}$ and $Center2_{B}$ respectively. 
	
	Without loss of generality, weights of the model are set to unity. Now, the following relationships hold on model at the time of presentation of test sample $TestPoint_{A}$.
	
	\begin{equation}\label{eq:18}
		dc1A = dc2B, 
	\end{equation}
	\begin{equation}\label{eq:19}
		dc2A=dc1B,
	\end{equation}
	\begin{equation}\label{eq:20}
		ac1A > ac1B > ac2B > ac2A, 
	\end{equation}
	\begin{multline}\label{eq:21}
		\phi_{c}(TestPoint_{A}, Center1_{A}) + \phi_{c}(TestPoint_{A}, Center2_{A}) \\ = \phi_{c}(TestPoint_{A}, Center1_{B}) + \phi_{c}(TestPoint_{A}, Center2_{B}).
	\end{multline}
	
	Let $\Psi$ is the discriminative power of a classifier. For Gaussian and cosine kernel classifer, their discriminative powers are respectively equivalent to:
	\begin{multline}\label{eq:22}
		\Psi_{g} = \phi_{g}(TestPoint_{A}, Center1_{A}) + \phi_{g}(TestPoint_{A}, Center2_{A}) \\- (\phi_{g}(TestPoint_{A}, Center1_{B}) + \phi_{g}(TestPoint_{A}, Center2_{B})),
	\end{multline}
	and
	\begin{multline}\label{eq:23}
	\Psi_{c} = \phi_{c}(TestPoint_{A}, Center1_{A}) + \phi_{c}(TestPoint_{A}, Center2_{A}) \\- (\phi_{c}(TestPoint_{A}, Center1_{B}) + \phi_{c}(TestPoint_{A}, Center2_{B})).
	\end{multline}
	
	Using \eqref{eq:18} and \eqref{eq:19}, we get:
	\begin{eqnarray}\label{eq:24}
		\Psi_{g} = 0,
	\end{eqnarray}
	similarly, using \eqref{eq:20} and \eqref{eq:21}, we get:
	\begin{eqnarray}\label{eq:25}
		\Psi_{c} = 0.
	\end{eqnarray}
	
	Since, both $\Psi_{g}$ and $\Psi_{c}$ are zero the probability that $TestPoint_{A}$ belong to $ClassA$ is equal to that of $ClassB$ i.e. equiprobable using either Gaussian or cosine classifier. The classification of $TestPoint_{A}$ is therefore solely dependent on the value of the bias.
	
	This lacking of correctly classifying a challenging cases such as $TestPoint_{A}$ persists even in RBF networks equipped with adaptive kernel fusion (Khan et al.) having global kernel weights as its discriminating power $\Psi_{a}$ for (Khan et al.) is defined as:
	
	\begin{eqnarray}\label{eq:26}
		\Psi_{a} = \alpha_{g}\Psi_{g}+\alpha_{c}\Psi_{c},
	\end{eqnarray}
	where $\alpha_{g} \in \mathbb{R}$ and $\alpha_{c} \in \mathbb{R}$ are (global) kernel coefficients of Gaussian and cosine kernels respectively. 
	
	Again for difficult cases such as $TestPoint_{A}$, it is verifiable that $\Psi_{a}=0$
	
	In contrast, the proposed method is not susceptible to such problems due to the novel concept of local weights (kernel coefficient) of each kernel. The discriminative power $\Psi_{r}$ of Co-RBFNN can be written as:
	
	\begin{multline}
	\Psi_{r} = \alpha_{Center1_{A},g}\phi_{g}(TestPoint_{A}, Center1_{A}) +\\ \alpha_{Center2_{A},g}\phi_{g}(TestPoint_{A}, Center2_{A}) \\
	+ \alpha_{Center1_{A},c}\phi_{c}(TestPoint_{A}, Center1_{A}) +\\ \alpha_{Center2_{A},c}\phi_{c}(TestPoint_{A}, Center2_{A}) \\
	-\big\{ \alpha_{Center1_{B},g}\phi_{g}(TestPoint_{A}, Center1_{B}) +\\ \alpha_{Center2_{B},g}\phi_{g}(TestPoint_{A}, Center2_{B}) \\
	+ \alpha_{Center1_{B},c}\phi_{c}(TestPoint_{A}, Center1_{B}) +\\ \alpha_{Center2_{B},c}\phi_{c}(TestPoint_{A}, Center2_{B})\big\},
	\end{multline}
	where $\alpha_{c,x} \in \mathbb{R}$ is the kernel coefficient for kernel of type $x$ and center $c$ such that $x \in {g,c}$ and $c \in {Center1_{A},  Center2_{A}, Center1_{B}, Center2_{B}}$ 
	
	It is evident that $\Psi_{r}\neq0$ as  $\alpha_{Center1_{A},g}\neq\alpha_{Center2_{A},g}$, $\alpha_{Center1_{A},c}\neq\alpha_{Center2_{A},c}$,
	$\alpha_{Center1_{B},g}\neq\alpha_{Center2_{B},g}$ and
	$\alpha_{Center1_{B},c}\neq\alpha_{Center2_{B},c}$ in general.

\subsection{Mean convergence analysis of our proposed model}\label{sec:convergence_co-rbf}
	{In this subsection, we mathematically prove that our proposed algorithm will effectively converge provided that we strategically set the learning rate $\eta$ less than $\lambda_{max}$, the maximum eigenvalue of the auto-correlation matrix $R$. We assume that, for the Wiener filter, the signal and (additive) noise are stationary linear stochastic processes with known spectral characteristics or known auto-correlation and cross-correlation \cite{haykin_adaptive_2014}.}

	The weight update rules of our proposed model i.e. \eqref{eq:simplified_weight_update_rule} and \eqref{eq:simplified_bias_update_rule} in the matrix form can be collectively rewritten as:
	\begin{equation}\label{eq:28}
		\bm{w}(n+1) = \bm{w}(n) + \eta \bm{\phi}(n)  e(n),
	\end{equation}
	where $\eta$ is the learning rate, $\bm{w}(n)$ is the weight vector of $n^{th}$ iteration and $e$ is the error between the desired and actual output signals i.e. 
	\begin{equation}\label{eq:29}
		e(n) = d(n) - y(n).
	\end{equation}
	  
		Let's define the vector $\bm{\Delta}_{opt}$ as the difference of our proposed model estimated weight vector $\bm{w}(n)$ with the optimal weight vector $\bm{w}_{opt}$:
		
		\begin{equation}\label{eq:30}
			\bm{\Delta}_{opt}(n) = \bm{w}(n) - \bm{w}_{opt},
		\end{equation}
		
		where optimal weight vector $\bm{w}_{opt}$ is that of Wiener filter obtained by solving the standard equation of Wiener filter  i.e.
		\begin{equation}\label{eq:31}
			\bm{P} - \bm{R}\bm{w}_{opt} = 0,
		\end{equation}
		where $\bm{P}$ is the cross-correlation matrix between input signal to $m$ hidden neurons (i.e. $\bm{\phi}$) and desired output $\bm{d}$, and $\bm{R}$ is the auto-correlation matrix of input signal to $m$ hidden neurons i.e. $\bm{\phi}$. Mathematically,
		\begin{align}
			\bm{R} &= E\Big(\bm{\phi}(n)\bm{\phi}^{T}(n)\Big), \label{eq:32} \\
			\bm{P} &= E\Big(\bm{\phi}(n)d\Big).
			\label{eq:33}	
		\end{align}
		
		Substituting the value of $\bm{e}$ from \eqref{eq:29} and subtracting $\bm{w}_{opt}$ from both sides of \eqref{eq:28}, we get:
		\begin{equation}
			\bm{\Delta}_{opt}(n+1) = \bm{\Delta}_{opt}(n) + \eta \bm{\phi}(n)\Big(d - y(n)\Big).
			\label{eq:34}
		\end{equation}
		Substituting the value of $y$ and $\bm{w}(n)$ from \eqref{eq:matrix-robust-rbf-nn} and \eqref{eq:30} respectively into \eqref{eq:28}, we get:
		\begin{equation}
			\bm{\Delta}_{opt}(n+1) = \bm{\Delta}_{opt}(n) + \eta \bm{\phi}(n) \Big(d-\bm{\phi}^{T}(n)(\bm{w}_{opt}+\bm{\Delta}_{opt}(n))\Big).
			\label{eq:35}
		\end{equation}
		
		Taking expectation on both sides of \eqref{eq:35} and rearranging few term, we obtain:
		\begin{multline}
		{E\Big(\bm{\Delta}_{opt}(n+1)\Big) = E\Big(\bm{\Delta}_{opt}(n)\Big) +  \eta E\Big(\bm{\phi}(n)d\Big) -} \\ {\eta E\Big(\bm{\phi}(n)\bm{\phi}^{T}(n) (\bm{w}_{opt}+\bm{\Delta}_{opt}(n)) \Big).}
		\end{multline}
		
		Further simplifying the above equation using \eqref{eq:31}, \eqref{eq:32} and \eqref{eq:33}, we get:
		\begin{equation}
			E\Big(\bm{\Delta}_{opt}(n+1)\Big) = E\Big(\bm{\Delta}_{opt}(n)\Big) - \eta E\Big(\bm{\phi}(n) \bm{\phi}^{T}(n) \bm{\Delta}_{opt}(n) \Big),
		\end{equation}
		
		After applying usual assumptions of Wiener filter \cite{haykin_adaptive_2014}, we obtain:
		\begin{equation}
		E\Big(\bm{\Delta}_{opt}(n+1)\Big) = \Big(I - \eta R \Big) E\Big(\bm{\Delta}_{opt}(n)\Big).
		\end{equation}
		
		Decomposing $R$ using singular value decomposition (SVD) and further simplification leads us to:
		\begin{equation}
			0 < \eta < \frac{1}{\lambda_{max}},
		\end{equation}
		where, $\lambda_{max}$ is the maximum eigenvalue of the autocorrelation matrix $R$.

\subsection{Mathematical analysis of the proposed model Co-RBFNN}\label{sec:analysis_co-rbf}
	{In this subsection, we mathematically prove that our proposed solution is superior to the adaptive kernel fusion \cite{khan2017novel}. We prove that the mean square error of our proposed solution is always less than that of the adaptive kernel fusion \cite{khan2017novel}. During this mathematical analysis, we made a usual assumption that the errors induced by the two models (i.e. our proposed solution and adaptive kernel fusion \cite{khan2017novel}) are zero mean Gaussian noise.}\footnote{Without loss of generality, the bias of the considered RBF models are assumed to be zero during the proofs of the following lemma and its two corollaries.} 

	\textbf{Lemma 1:} Our proposed model has following relationship with adaptive kernel fusion (Khan et al.) model \cite{khan2017novel}
	\begin{equation}\label{eq:39}
	y_{d} = y_{a} + e_{x},
	\end{equation}  
	where, $y_{d}$ and $y_{a}$ are the estimated responses of our proposed model and adaptive kernel fusion \cite{khan2017novel} respectively and $e_{x}$ is the noise. Mathematically, the estimated responses of the two models $y_{a}$ and $y_{d}$ respectively are defined as:
	
	\begin{equation}\label{eq:40}
	y_{a} = \alpha \bm{w}^{T}\bm{\phi_{g}} + (1 - \alpha) \bm{w}^{T}\bm{\phi_{c}},
	\end{equation}
	and,
	\begin{equation}\label{eq:41}
	y_{d} = \bm{w}_{g}^{T} \bm{\phi}_{g}(\bm{x}) +\bm{w}_{c}^{T} \bm{\phi}_{c}(\bm{x}),
	\end{equation} 
	where $\bm{w}_{g}$ and $\bm{w}_{c}$ are Gaussian and cosine weight vectors of our proposed model respectively and, $\bm{w}$ and $\alpha$ are the weight vector and multi-kernel coefficient of adaptive kernel fusion \cite{khan2017novel} respectively.
	
	\textbf{Prove:} Consider our proposed model that estimates the desired response by minimizing the least square error i.e.
	\begin{equation}\label{eq:42}
		d = y_{d} + e,
	\end{equation}
	where, $d$ is the desired response vector, $y_{d}$ is the estimated response of our proposed model and $e \in \mathcal{N}(0,\sigma)$ is the Gaussian noise of the proposed model.
	 
	Further, the following relationships hold among weight vectors $\bm{w}$, $\bm{w}_{g}$ and $\bm{w}_{c}$:
	
	\begin{equation}\label{eq:43}
		\bm{w}_{g} = \alpha \bm{w} + \bm{e}_{g},
	\end{equation}
	\begin{equation}\label{eq:44}
	\bm{w}_{c} = (1 - \alpha) \bm{w} + \bm{e}_{c},
	\end{equation}
	where $\bm{e}_{g} \in \mathcal{N}(0,\sigma_{g})$ and $\bm{e}_{c} \in \mathcal{N}(0,\sigma_{c})$ are Gaussian noises and $\alpha$ is the kernel coefficient of multi-kernel as defined in adaptive kernel fusion \cite{khan2017novel}.
	
	By adding \eqref{eq:43} and \eqref{eq:44}, we get another relation i.e.
	\begin{equation}\label{eq:45}
	\bm{w}_{g} + \bm{w}_{c} = \bm{w} + \bm{e}_{g} + \bm{e}_{c}.
	\end{equation}
	Adding and subtracting the term $\bm{w_{g}}^{T}\bm{\phi_{c}}(\bm{x})$ on R.H.S of \eqref{eq:39}, substituting the value of $y_{d}$ from \eqref{eq:41} and simplifying, we get:
	\begin{eqnarray}\label{eq:46}
		d = \bm{w_{g}}^{T}(\bm{\phi_{g}}(\bm{x}) - \bm{\phi_{c}}(\bm{x})) + (\bm{w_{g}} + \bm{w_{c}})^{T}\bm{\phi_{c}}(\bm{x}) + \bm{e}.
	\end{eqnarray}
	
	After substituting the value of $\bm{w_{g}}$ from \eqref{eq:43} and that of $(\bm{w_{g}} + \bm{w_{c}})$ from $\eqref{eq:45}$ into \eqref{eq:46} and simplifying, we obtain:
	\begin{eqnarray}\label{eq:47}
		d = \alpha \bm{w}^{T} \bm{\phi_{g}} + (1 - \alpha) \bm{w}^{T} \bm{\phi_{c}} + \bm{e}_{g}^{T}\bm{\phi_{g}}(\bm{x}) + \bm{e}_{c}^{T}\bm{\phi_{c}}(\bm{x}) + \bm{e}.
	\end{eqnarray} 
	
	After substituting the value of $\alpha \bm{w}^{T} \bm{\phi_{g}} + (1 - \alpha) \bm{w}^{T} \bm{\phi_{c}}$ from \eqref{eq:40}, we obtain:
	\begin{eqnarray}\label{eq:48}
	d =  y_{a} + \bm{e}_{g}^{T}\bm{\phi_{g}}(\bm{x}) + \bm{e}_{c}^{T}\bm{\phi_{c}}(\bm{x}) + \bm{e}.
	\end{eqnarray}
	  
	Let the error term $\bm{e}_{g}^{T}\bm{\phi_{g}}(\bm{x}) + \bm{e}_{c}^{T}\bm{\phi_{c}}(\bm{x})$ be represented as $\bm{e}_{x}$, \eqref{eq:48} becomes:
	
	\begin{eqnarray}\label{eq:49}
	d = y_{a} + \bm{e}_{x} + \bm{e},
	\end{eqnarray}
	substituting the value of $d$ from \eqref{eq:42} into \eqref{eq:49} and simplifying, we get:
	\begin{eqnarray}\label{eq:50}
	y_{d} = y_{a} + e_{x}, \qquad\text{Q.E.D}
	\end{eqnarray}
	
	\textbf{Corollary 1:} The error term $e_{x}$ is mean zero Gaussian noise i.e. $e_{x} \in \mathcal{N}(0,\sigma_{x})$.
	
	\textbf{Prove:}
	Since adaptive kernel fusion \cite{khan2017novel} estimates the desired response $d$ by minimizing the least square error. Therefore, it is mathematically definable as:
	\begin{eqnarray}\label{eq:51}
	d = y_{a} + e_{a},
	\end{eqnarray}
	where, $y_{a}$ is the estimated response and $e_{a} \in \mathcal{N}(0,\sigma_{a})$ is the Gaussian noise of the model respectively and $d$ is the desired response vector.
	
	Substituting the value of $d$ from \eqref{eq:49} into \eqref{eq:51} and simplifying, we get:
	\begin{eqnarray}\label{eq:52}
	e_{x} = e_{a} - e.
	\end{eqnarray}
	
	Since, $e_{x}$ is the difference of two zero mean Gaussian noises i.e. $e$ and $e_{a}$,  $e_{x}$ is also a zero mean Gaussian noise i.e. $e_{x} \in \mathcal{N}(0,\sigma_{x})$, hence proved.
	
	\textbf{Corollary 2:} Mean squared error of adaptive kernel fusion (Khan et al.) model \cite{khan2017novel} $\|e_{a}\|_{2}^{2}$ is always greater than or equal to that of our proposed model $\|e_{a}\|_{2}^{2}$ i.e.
	 
	\begin{eqnarray}\label{eq:53}
	\|e_{a}\|_{2}^{2} \geq \|e\|_{2}^{2}.
	\end{eqnarray} 
	
	\textbf{Prove:}
	Substituting the value of $d$ from \eqref{eq:49} into \eqref{eq:51} and simplifying, we get:
	\begin{eqnarray}\label{eq:54}
	e_{a} = e + e_{x},
	\end{eqnarray}
	
	Since, $e_{a} \in \mathcal{N}(0,\sigma_{a})$ is the sum of two mean zero Gaussian noises i.e. $e \in \mathcal{N}(0,\sigma)$ and $e_{x} \in \mathcal{N}(0,\sigma_{x})$.  Hence,
	\begin{eqnarray}\label{eq:55}
	\sigma_{a}^{2} = \sigma^{2} + \sigma_{x}^{2}.
	\end{eqnarray}
	This lead us to:
	\begin{equation*}
	\|e_{a}\|_{2}^{2} = \|e\|_{2}^{2} + \|e_{x}\|_{2}^{2},
	\end{equation*}
	so,
	\begin{equation*}
	\|e_{a}\|_{2}^{2} \geq \|e\|_{2}^{2},
	\end{equation*}
	hence, proved.      

\section{Experimental results}
\label{sec:results}

In this section, we compare the performance of our proposed solution against two state-of-the-art multi-kernel radial basis function neural network algorithms namely manually fused multi-kernel proposed by Aftab et.al \cite{aftab_novel_2014} and adaptively fused multi-kernel proposed by Khan et.al in \cite{khan2017novel}. All three algorithms are tested on pattern classification, system identification and function approximation problems for standard performance measures. All tests are preformed using Matlab R2017b on Intel CORE i5-2540M CPU @2.60GHz 4GB RAM. Results are averaged over 100 independent random runs.

\subsection{Pattern classification}
{Pattern classification has several applications in security, industry, medicine and defense. Examples include iris identification, speaker identification, fingerprint identification,  statistical pattern recognition of seismic data, and automatic medical diagnosis.}

A well known Iris flower dataset \cite{fisher_use_1936} is selected for pattern classification problem.  The dataset consist of three classes (flower species). Each class has 50 samples and four attributes i.e. sepal length, sepal width, petal length, and petal width. Forty samples of each class are randomly selected for training where as remaining ten samples of each class are used for testing.

The three RBF networks are trained with the following specifications. $16$ neurons are used with kernel centers selected using subtractive clustering \cite{pal_mountain_2000} with influence factor $0.2$. Gaussian kernel width is set to unity. Learning rate is $5\times10^{-3}$. The weights as well as bias are initialized randomly.

\begin{figure}[ht!] 
		\centering
		\includegraphics[width=0.7\textwidth]{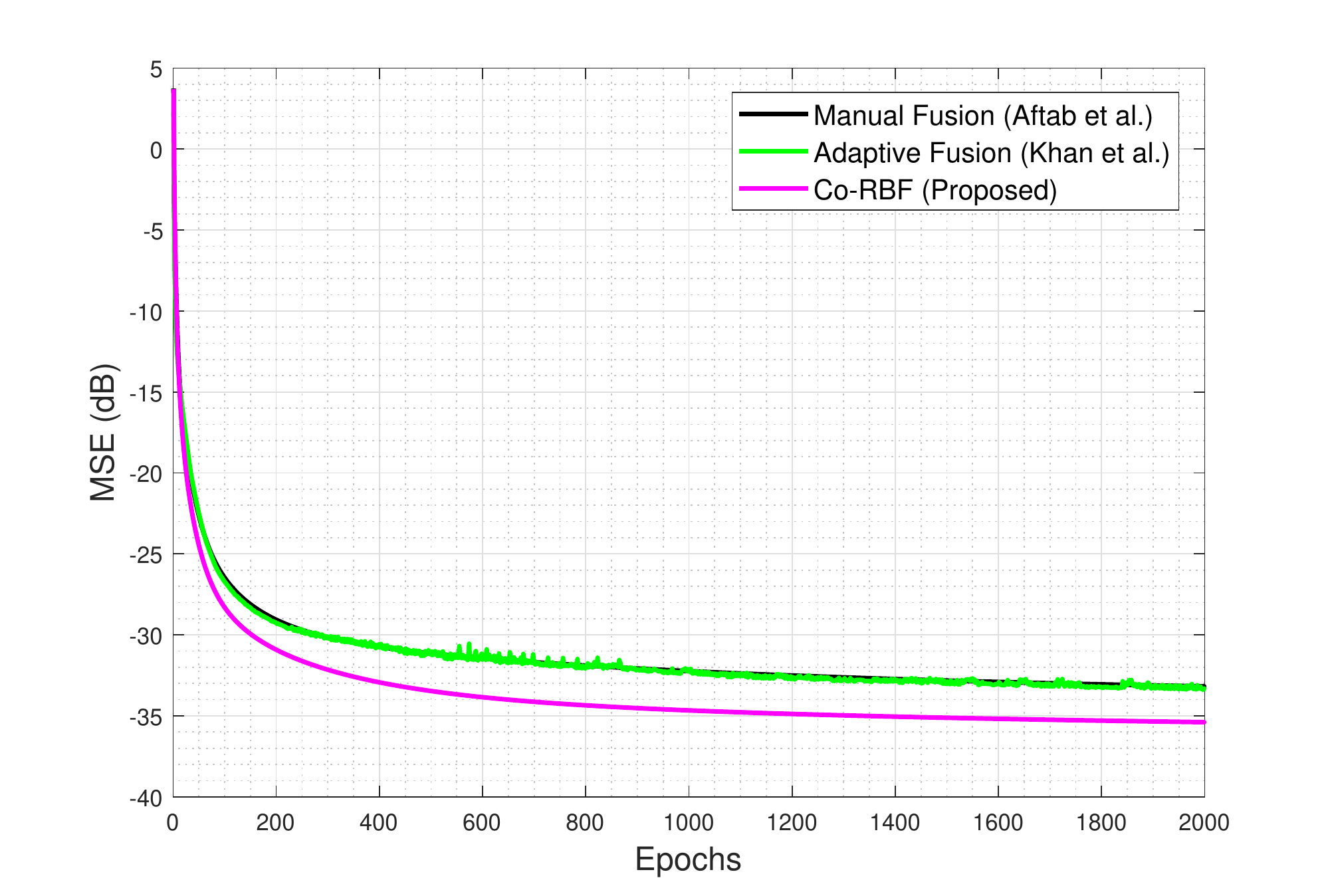}
		\caption{MSE curves of different RBF algorithms on Iris Flowers dataset. \label{fig:mse_patt_class_iris}} 
	\end{figure}

Fig.~\ref{fig:mse_patt_class_iris} shows MSE curves obtained during training. It is evident that our proposed architecture requires only $160$ epochs to achieve mean squared error of $-30.17$ dB whereas the other two algorithms require at least $240$ epochs to reach the same MSE. Moreover, the proposed architecture settles on an MSE of $-35.39$ dB after $2000$ epoch whereas the other two algorithms achieve a worse error of $-33.33$ dB after same number of epochs. Hence, our proposed architecture outperforms other two state-of-the-art techniques both in term of rate of convergence and steady-state error.

\begin{figure}[ht!] 
		\centering
		\includegraphics[width=0.7\textwidth]{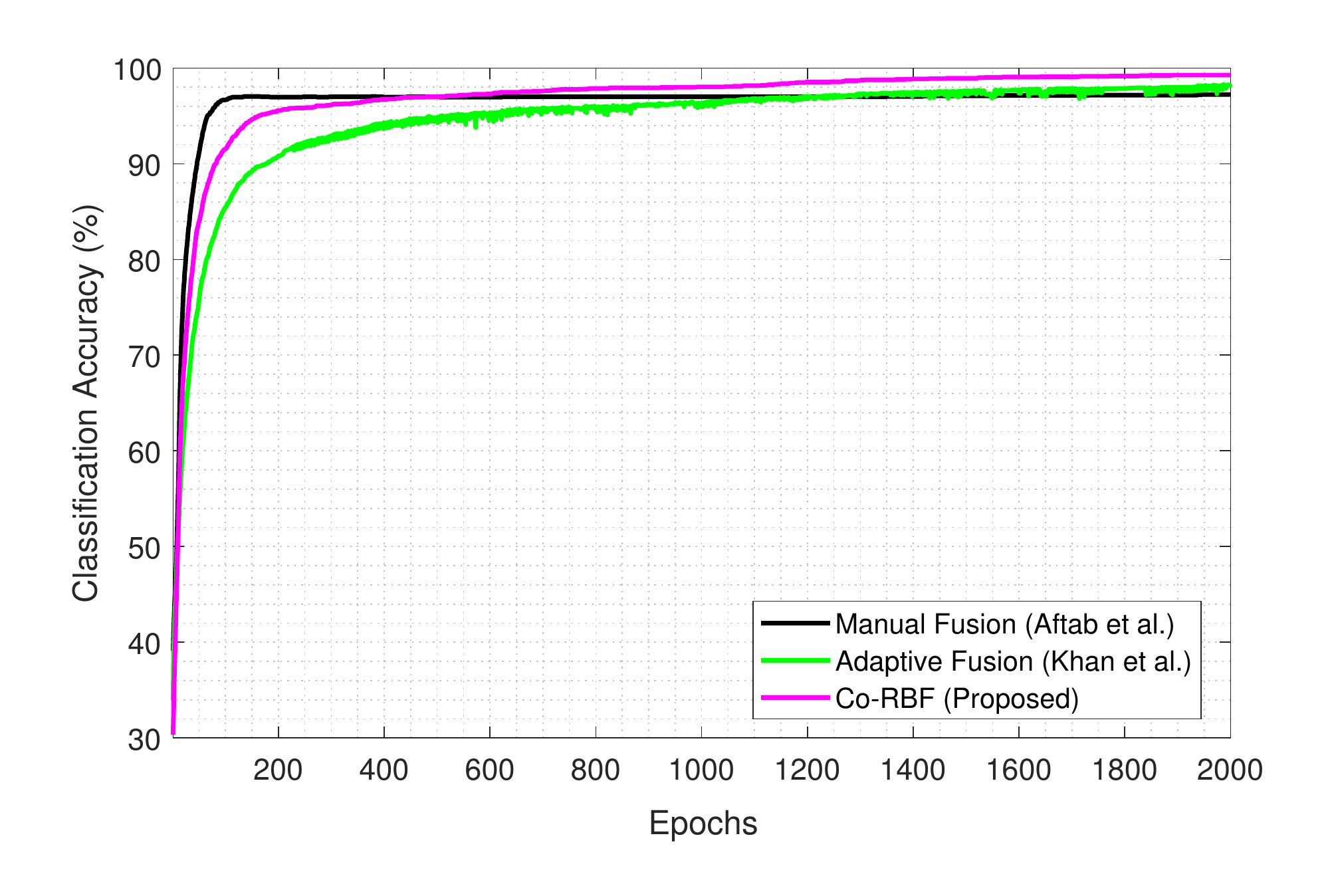}
		\caption{Training accuracy curves of different RBF algorithms on Iris Flowers Dataset. \label{fig:training_accuracy_class_iris}} 
	\end{figure}

\begin{figure}[ht!] 
		\centering
		\includegraphics[width=0.7\textwidth]{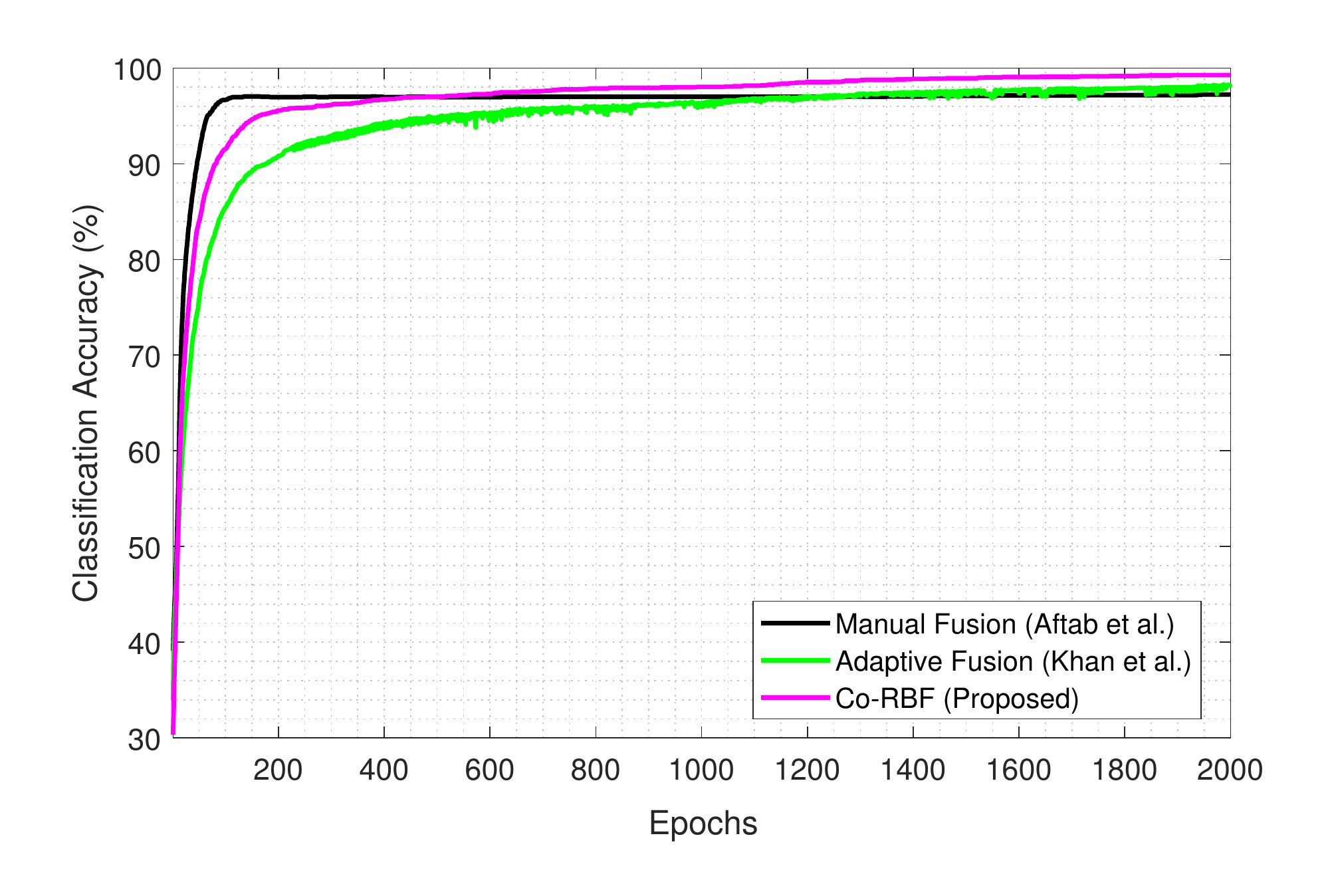}
		\caption{Testing accuracy curves of different RBF algorithms on Iris Flowers dataset. \label{fig:testing_accuracy_class_iris}} 
	\end{figure}

Classification accuracy achieved by different RBF algorithms on the given dataset is shown in Table \ref{tb:accuracy_iris}. During the training phase, the proposed architecture showed accuracy of $98.35\%$ that is $0.64\%$ higher than that manual kernel fusion \cite{aftab_novel_2014} but $0.24\%$ less compared to the adaptive kernel fusion \cite{khan2017novel} that attain the accuracy of $98.59\%$. However, our proposed approach attained the best testing accuracy of $99.13\%$ comparing to $97.00\%$ that of manual kernel fusion \cite{aftab_novel_2014} and $98.50\%$ that of adaptive kernel fusion \cite{khan2017novel}. It established that the proposed architecture is significantly tolerable to over-fitting. Moreover, our architecture is even not susceptible to the initial weights (and the bias) as it exhibited the lowest standard deviation of $0.12\%$ (on the training data) and the second lowest standard deviation of $1.47\%$ (on the test data). Fig. ~\ref{fig:training_accuracy_class_iris} and Fig. ~\ref{fig:testing_accuracy_class_iris} show the training and testing accuracy curves of the three approaches respectively. Our proposed architecture exhibited better training accuracy from the start thus achieved the training accuracy of $95.67\%$ at $100$ epoch whereas the other two algorithm achieved $92.84\%$ only at $100$ epoch. On testing data, the manual kernel fusion \cite{aftab_novel_2014} initially exhibited the best accuracy precisely $96.5\%$ at $100$. But, our proposed approach became the best at $600$ epoch and marked the best steady-state accuracy of $99.27\%$ at $2000$ epoch comparing to that $98.27\%$ by adaptive kernel fusion \cite{khan2017novel} and $97.23\%$ by manual kernel fusion \cite{aftab_novel_2014}.

	\begin{table}[!h]
		\caption{Classification accuracy (in \%) of Iris Flowers dataset obtained by different RBF algorithms}
		\label{tb:accuracy_iris}
		\begin{center}
			\begin{tabular}{|c|r|r|}
			\hline
			\multicolumn{ 1}{|c|}{Method} & \multicolumn{ 1}{c|}{Training} & \multicolumn{ 1}{c|}{Testing} \\
			\multicolumn{ 1}{|c|}{} &       \multicolumn{ 1}{c|}{mean$\pm$std} &  \multicolumn{1}{c|}{mean$\pm$std}\\
			\hline
			Manual Fusion (Aftab et al.) &      	97.71$\pm$0.61 &			97.00$\pm$1.01 \\
			\hline
			Adaptive Fusion (Khan et al.) & \textbf{98.59}$\pm$1.12 &			98.50$\pm$4.68 \\
			\hline
			Co-RBF (Proposed)&      98.35$\pm$0.12 &	\textbf{99.13}$\pm$1.47 \\
			\hline
			\end{tabular} 	
		\end{center}
	\end{table} 

Sensitivity and specificity are also two important performance metric to analyze a classifier for its biasedness of a classifier. Sensitivity and specificity of different algorithms are tabulated in Table \ref{tb:sensitivity_iris} and Table \ref{tb:specificity_iris} respectively. Our proposed algorithm exhibits the best sensitivity of $97.50\%$ and $100\%$ on Versicolor and Setosa classes respectively during training and that of $100\%$ and $100\%$ on Virginica and Versicolor classes respectively in testing phases. Moreover, the sensitivity obtained by the proposed algorithm for all three classes are very close to each other in the range of $0\%$ to $0.35\%$ in testing phase showing unbiasedness of the proposed method. 

	\begin{table}[!h]
		\caption{Average classification sensitivity (in \%) of Iris Flowers obtained by different RBF algorithms after training for 2000 epochs}
		\label{tb:sensitivity_iris}
		\begin{center}
			\begin{tabular}{|c||c||c|c|c|}
				\hline
				\multirow{2}{*}{Architecture}                                                        & \multirow{2}{*}{Phase} & Virginica & Versicolor & Setosa \\ \cline{3-5}
				& & mean$\pm$std & mean$\pm$std & mean$\pm$std \\ \hline
				\multirow{2}{*}{Manual Fusion (Aftab et al.)} 		& Training & 97.10$\pm$1.58 & 96.03$\pm$1.24 & 											100$\pm$0.00 \\ \cline{2-5}
				& Testing & 100$\pm$0.00 & 100$\pm$0.00 & 91.00$\pm$3.02 \\ \hline
				\multirow{2}{*}{Adaptive Fusion (Khan et al.)} 	& Training & \textbf{98.65}$\pm$1.644 & 97.13$\pm$2.11 & 											100$\pm$0.00 \\ \cline{2-5}
				& Testing  & 100$\pm$0.00 & 97.40$\pm$13.83 & \textbf{98.10}$\pm$3.94 \\ \hline
				\multirow{2}{*}{Co-RBF (Proposed)} 	& Training & 97.55$\pm$0.35 & \textbf{97.50}$\pm$0.00 & 											\textbf{100}$\pm$0.00 \\ \cline{2-5}
				& Testing  & \textbf{100}$\pm$0.00 &	\textbf{100}$\pm$0.00 & 97.40$\pm$4.41 \\ \hline   
			\end{tabular}
		\end{center}
	\end{table}

During the training phase, our proposed algorithm shows the best specificity of $98.75\%$ and $100\%$ on Versicolor and Setosa classes respectively. Whereas, it achieved the average specificity of $98.75$ on Versicolor class which is the second best specificity (i.e. $0.55\%$ less than that of the best specificity of $99.33\%$ reached by adaptive kernel fusion \cite{khan2017novel}) on that class. Specificity results of testing phase are also very similar. Our algorithm attained the specificity of $100\%$ on both Versicolor and Setosa classes. However, it achieved the specificity of $98.70\%$ on Versicolor class which is the second best specificity on that class, $0.35\%$ less than the best ($99.05\%$) attained by adaptive kernel fusion \cite{khan2017novel}.

	\begin{table}[!h]
		\caption{Average classification specificity (in \%) of Iris Flowers obtained by different RBF algorithms after training for 2000 epochs}
		\label{tb:specificity_iris}
		\begin{center}
			\begin{tabular}{|c||c||c|c|c|}
				\hline
				\multirow{2}{*}{Architecture}                                                        & \multirow{2}{*}{Phase} & Virginica & Versicolor & Setosa \\ \cline{3-5}
				& & mean$\pm$std & mean$\pm$std & mean$\pm$std \\ \hline
				\multirow{2}{*}{Manual Fusion (Aftab et al.)} 		& Training & 98.01$\pm$0.62 & 98.55$\pm$0.79 &														100$\pm$0.00 \\ \cline{2-5}
															& Testing & 100$\pm$0.00 & 95.50$\pm$1.51 & 			100$\pm$0.00 \\ \hline
				\multirow{2}{*}{Adaptive Fusion (Khan et al.)} 	& Training & 98.56$\pm$1.06		 & 														\textbf{99.33}$\pm$0.82 & 100$\pm$0.00 \\ \cline{2-5}
															& Testing  & 98.70$\pm$6.91 & \textbf{99.05}$\pm$1.97 & 			100$\pm$0.00 \\ \hline
				\multirow{2}{*}{Co-RBF (Proposed)} 	& Training & \textbf{98.75}$\pm$0.00 & 98.78$\pm$0.18 & 														\textbf{100}$\pm$0.00 \\ \cline{2-5}
															& Testing  & \textbf{100}$\pm$0.00 & 98.70$\pm$2.20 & 			\textbf{100}$\pm$0.00  \\ \hline   
			\end{tabular} 
		\end{center}
	\end{table}

Table~\ref{tb:youden_index_iris} is showing Youden index of different algorithms on Iris Flowers dataset. It is a popular index used to quantified the overall capacity of the model for pattern classification. During the training phase, adaptive kernel fusion\cite{khan2017novel} attained the best indices of $0.9721$, $0.9646$ and $1.0000$ for Virginica, Versicolor and Setosa classes respectively. Followed by our algorithm with indices of $0.9630$ ($0.0091$ less than the best), $0.9628$ ($0.0018$ less than the best) and $1.0000$ for Virginica, Versicolor and Setosa classes respectively. Manual kernel fusion\cite{aftab_novel_2014} is in the last with indices of $0.9511$, $0.9458$ and $1.0000$ for Virginica, Versicolor and Setosa classes respectively.

During testing phase, our algorithm achieved the best Youden indices of $1.0000$ and $0.9870$ for classes  Virginica and Versicolor respectively. However, it attained the second best Youden index of $0.9740$ on Setosa class (i.e. $0.0070$ less than $0.9810$ the best Youden index reached by adaptive kernel fusion\cite{khan2017novel}). In the light of our simulation results of Virginica and Versicolor classes, adaptive kernel fusion\cite{khan2017novel} is the second best (with Youden indices of $0.9870$ and $0.9745$ for Virginica and Versicolor classes respectively) and manual kernel fusion\cite{aftab_novel_2014} is the worst (with Youden indices of $1.0000$ and $0.9550$ for Virginica and Versicolor classes respectively) in term of Youden index during testing phase.

	\begin{table}[!h]
		\caption{Average Youden index of Iris Flowers obtained by different RBF algorithms after training for 2000 epochs}
		\label{tb:youden_index_iris}
		\begin{center}
			\begin{tabular}{|c||c||c|c|c|}
				\hline
				Architecture& Phase & Virginica & Versicolor & Setosa \\ \hline
				\multirow{2}{*}{Manual Fusion (Aftab et al.)} & Training & $0.9511$ & $0.9458$ & $1.0000$ \\ \cline{2-5}
				& Testing & 1.0000 & 0.9550 & 0.9100 \\ \hline				
				\multirow{2}{*}{Adaptive Fusion (Khan et al.)} 	& Training & \textbf{0.9721} & \textbf{0.9646} & 1.0000 \\ \cline{2-5}
				& Testing  & 0.9870 & 0.9745 & \textbf{0.9810} \\ \hline
				\multirow{2}{*}{Co-RBF (Proposed)} 	& Training & 0.9630 & 0.9628 & \textbf{1.0000} \\ \cline{2-5}
				& Testing  & \textbf{1.0000} & \textbf{0.9870} & 0.9740  \\ \hline   
			\end{tabular} 
		\end{center}
	\end{table}
\subsection{Function approximation problem}
{Function approximation is a way to describe the behavior of complicated functions using available observations from the domain through ensembles of simpler functions. It has special importance in several research domains like dynamic system modeling, nonlinear complex-valued signal processing, and biological activity modeling etc \cite{877615,khan2019universal,yoon2018efficient}.}

For the function approximation problem, we consider the following non linear function defined as: 
	\begin{equation}
	f(x_{1}, x_{2})=e^{(x_{1}^{2} - x_{1}^{2})}, \quad \forall \; -1 \leq x_{1} \leq1 \; \text{and} \; -1 \leq x_{2} \leq 1,
	\end{equation}
For training phase, $x_{1}$ and $x_{2}$ were selected over the interval $[-1,1]$ with sampling spacing of $0.2$. Whereas for the testing phase, $x_{1}$ and $x_{2}$ were selected over the interval $[-0.9,0.9]$ at the same rate. Hence, $121$ and $100$ samples were used for training and testing respectively.   

All the RBF algorithms were initialized with the following specifications. Learning rate was set to $1\times10^{-3}$ and the Gaussian kernel spread was taken to be unity. All $121$ hidden neurons were configured by selecting training samples as centers for the kernel. Weights and bias were initialized randomly for every run. 

\begin{figure}[ht!] 
		\centering
		\includegraphics[width=0.7\textwidth]{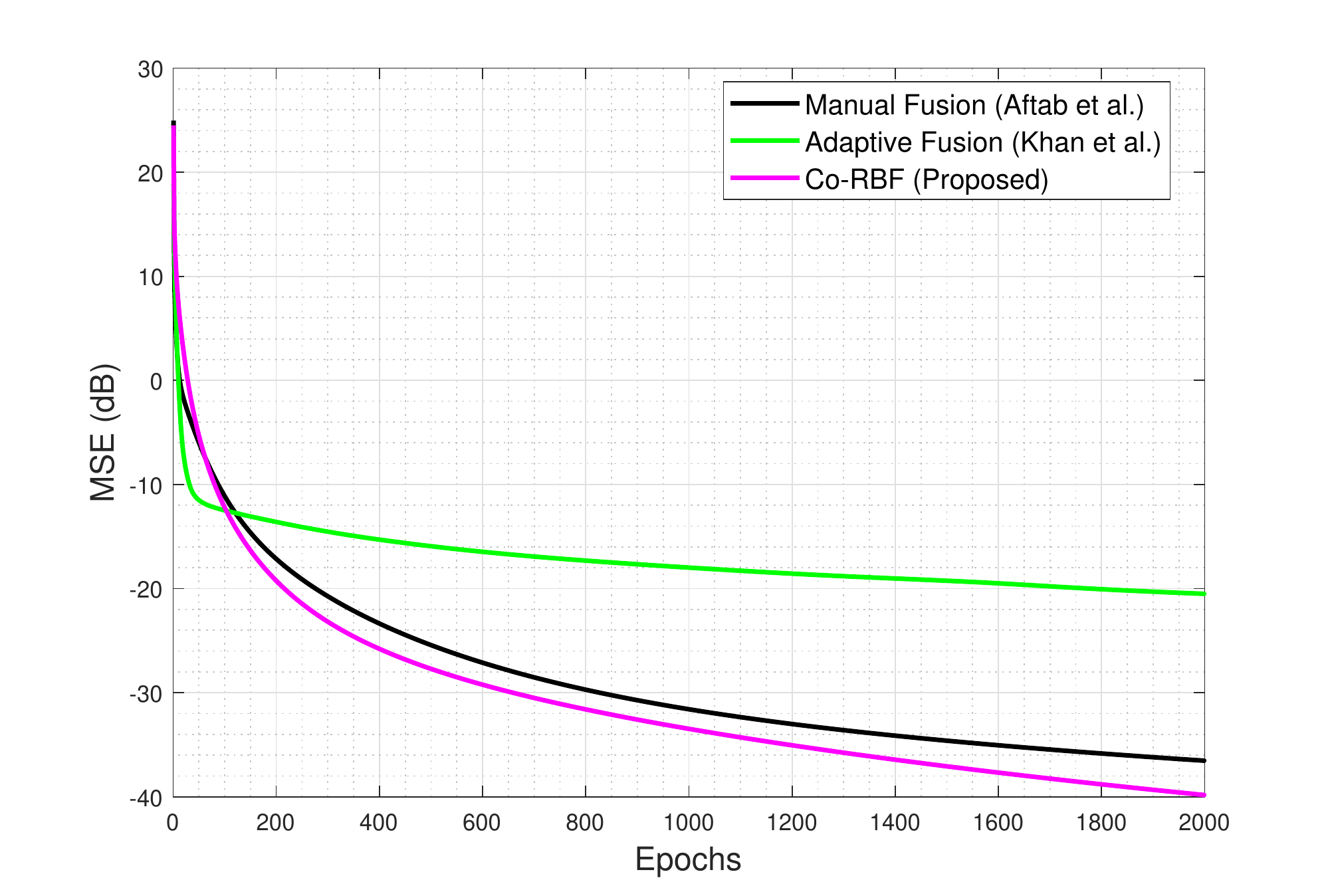}
		\caption{MSE curves of different RBF algorithms on function approximation problem. \label{fig:mse_fun_appox}} 
	\end{figure}

MSE curves of different RBF algorithms during training are shown in Fig.~\ref{fig:mse_fun_appox}. Adaptive kernel fusion architecture \cite{khan2017novel} showed the highest convergence rate for first $50$ epochs but then got stuck in a local minima and achieved the higher error of $-20.5$ db at $2000$ epochs. In contrast, our proposed architecture showed moderate but consistent convergence rate thus achieved the minimum error $-39.83$ dB at $2000$ epochs. Moreover, manual kernel fusion architecture \cite{aftab_novel_2014} exhibited moderate final convergence by attaining the error of $-36.53$ db at $2000$ epochs. 

\begin{figure}[ht!] 
		\centering
		\includegraphics[width=0.7\textwidth]{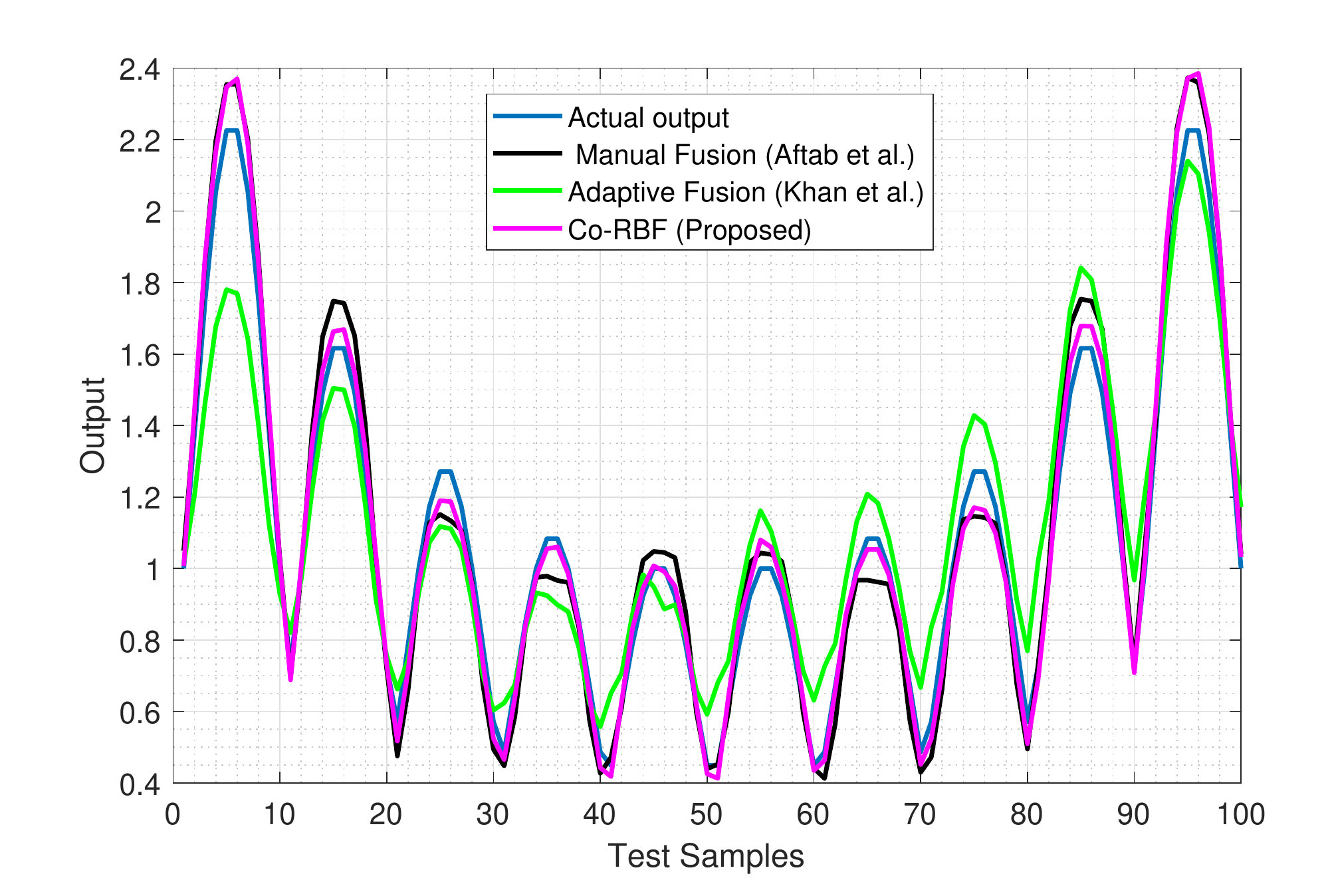}
		\caption{Predicted output of different RBF algorithms on test data of function approximation problem. \label{fig:output_fun_appox}} 
	\end{figure}

\begin{figure}[ht!] 
		\centering
		\includegraphics[width=0.7\textwidth]{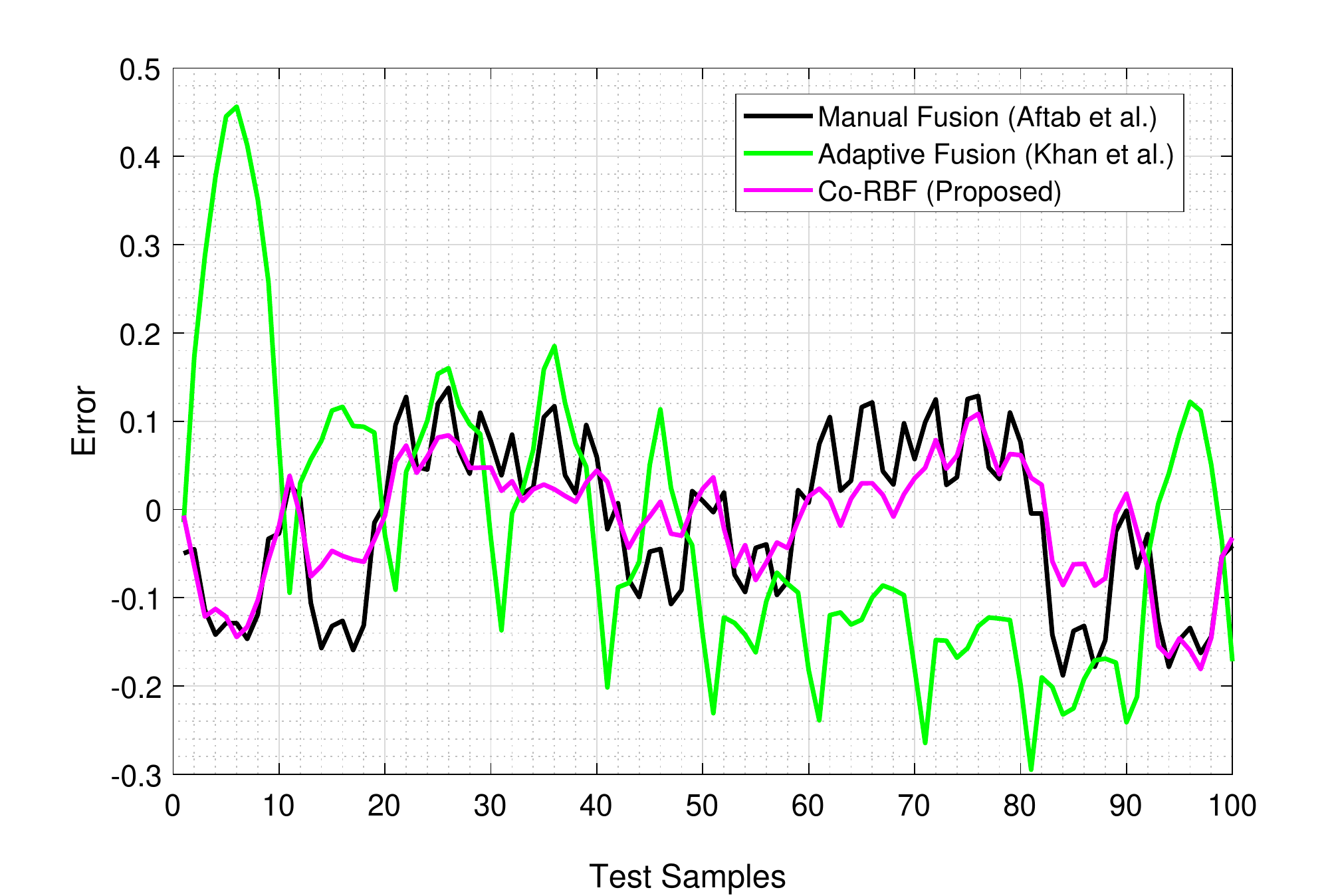}
		
		\caption{Instantaneous error of different RBF algorithms on test Data of function approximation problem. \label{fig:inst_err_fun_appox}} 
	\end{figure}

Instantaneous error of our proposed architecture is well bounded between $-0.1$ and $0.1$ whereas that of manual kernel fusion \cite{aftab_novel_2014} is bounded between $-0.15$ and $0.15$ and that of Adaptive kernel fusion \cite{khan2017novel} is bounded between $4.5$ and $-3.0$ as depicted in \ref{fig:inst_err_fun_appox}. Hence,  Adaptive kernel fusion \cite{khan2017novel} is the worst in term of instantaneous error among all the three algorithms. As the result, the predicted output of our proposed architecture mapped the actual output in the best manner as showed in Fig.~\ref{fig:output_fun_appox}.
  
	\begin{figure}[ht!] 
		\centering
		\includegraphics[width=0.7\textwidth]{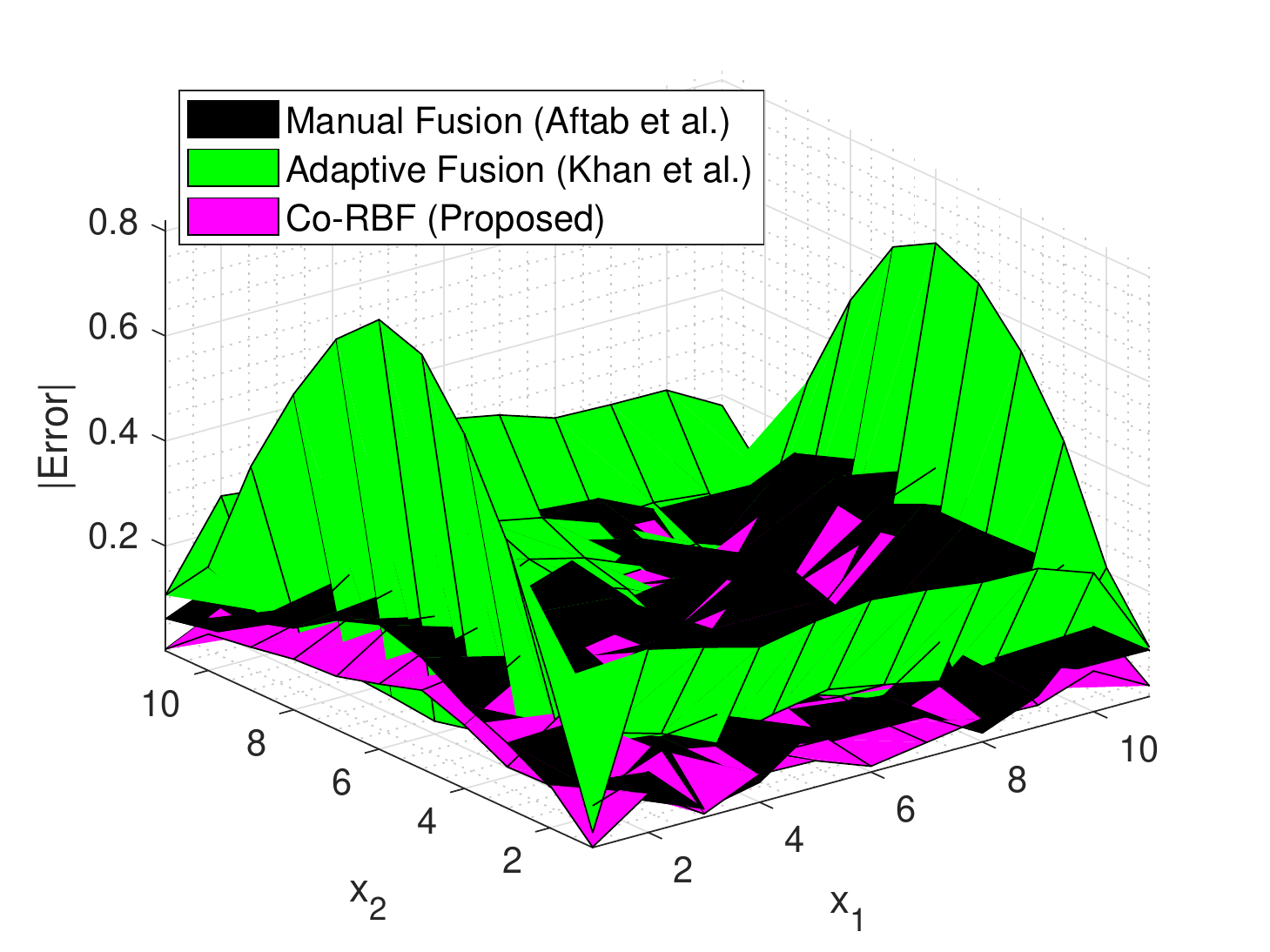}
		\caption{Error surfaces of different RBF algorithms on train data of function approximation Problem \label{fig:train_err_surf_fun_appox}} 
	\end{figure}
	
	\begin{figure}[ht!] 
		\centering
		\includegraphics[width=0.7\textwidth]{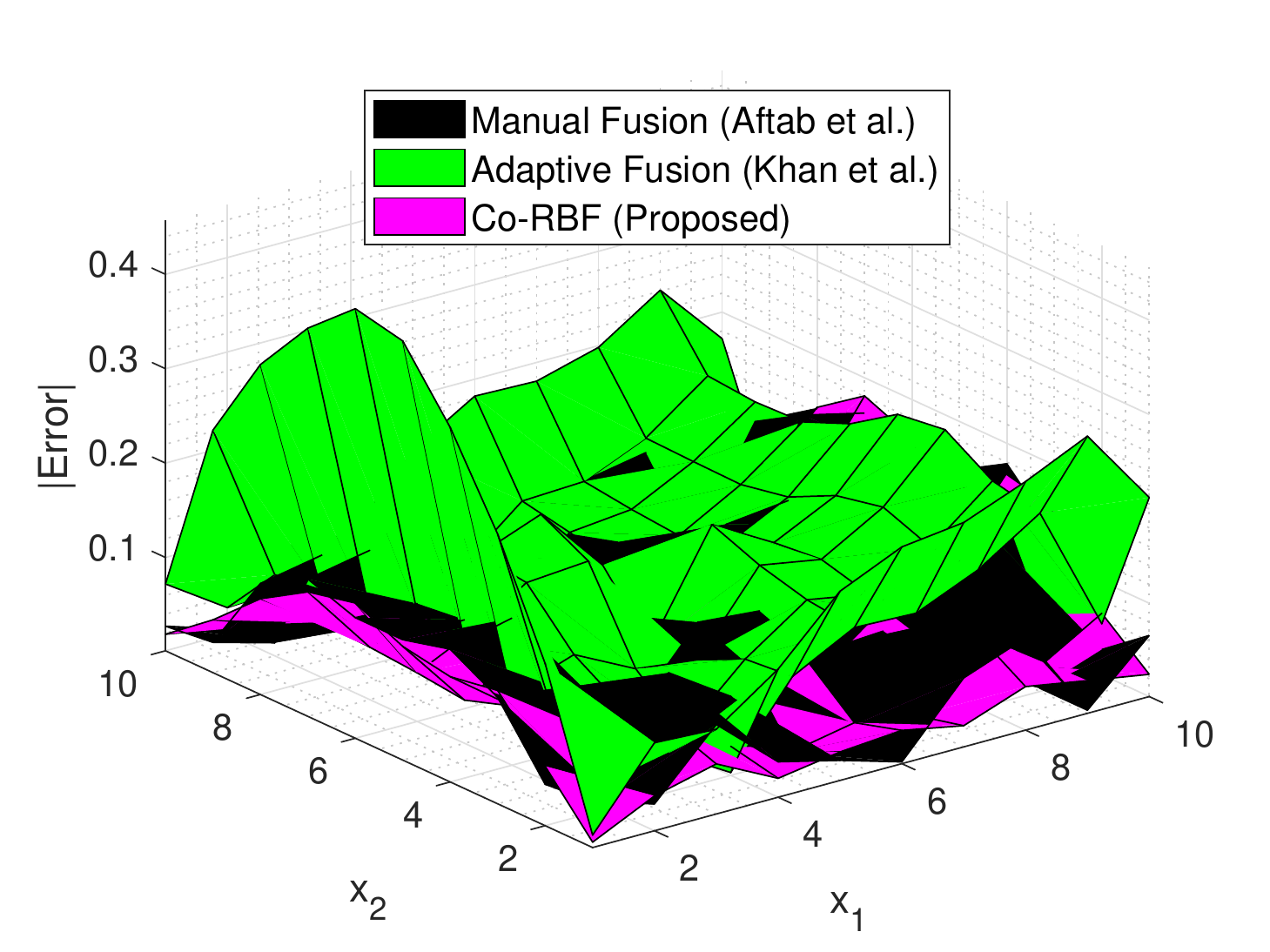}
		\caption{Error surfaces of different RBF algorithms on test data of function approximation problem. \label{fig:test_err_surf_fun_appox}} 
	\end{figure}

Fig.~\ref{fig:train_err_surf_fun_appox} and \ref{fig:test_err_surf_fun_appox} are showing the error surfaces of different RBF algorithms on training and testing data. Error surface of Adaptive kernel fusion \cite{khan2017novel} is quite spiky for both the training and testing data i.e. bounded between $4.5$ and $-3.0$ (training data) and $8.0$ and $-3.5$ (testing data) respectively. It indicates that the algorithm poorly approximated the given function. In contrast, error surfaces of our proposed architecture are very flat bounded between $1.0$ and $-1.0$ in case training data and that $-0.12$ and $-0.14$ in case of testing data. This indicates that given function is well approximated by Co-RBFNN. Manual kernel fusion is moderately spiky with error bound of $(-0.15,0.15)$ for training data and that of $(-0.22,0.13)$ for testing data. Thus, its ability of function approximation of the given function is average.    

\subsection{Nonlinear System Identification}
\begin{figure}[ht!]
		\begin{center}
			\centering 
			\includegraphics[width=3.5in,keepaspectratio]{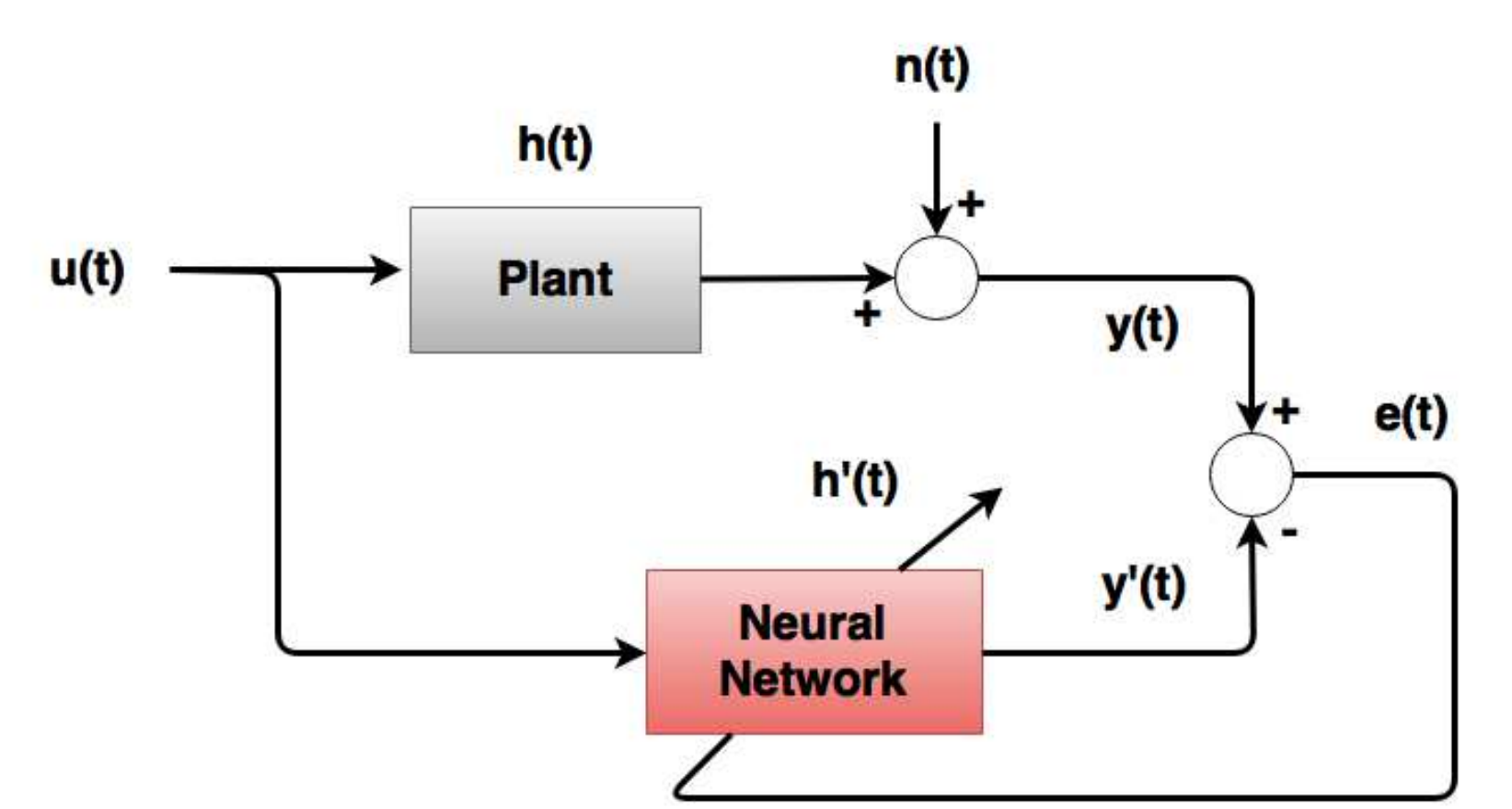} 
		\end{center}
		\caption{Nonlinear system identification using RBF neural network.}
		\label{fig:non_linear_sys_ident}
	\end{figure}

{System identification/nonlinear system identification is a systematic approach to build mathematical models of dynamic systems using measurements of only the system's input and output signals. It has several applications in diverse fields ranging from wireless communication systems \cite{ahmad2017fclms,sadiq2019enhanced,khan2017flmf} to geo localization of mines \cite{nerguizian2006geolocation} etc. It is considered to be a highly challenging research problem in the domain of signal processing and can be effectively addressed using neural networks \cite{khan2018novel}. Fig.~\ref{fig:non_linear_sys_ident} depicts a general systematic approach used by the RBF neural networks for this purpose. For the evaluation of the proposed architecture, we consider a first order non linear system defined by the following equation:}
 
{
\begin{equation}\label{eq:sys_id_ex_1}
		y_t=2u_{(t)} - 0.5u_{(t-1)} -0.1 u_{(t-2)} -0.7 (cos(3 u_{(t)}) + \mathrm{e}^{-|u_{(t)}|}),
	\end{equation}      
where, $u_{t}$ and $y_{t}$ are the system input and output respectively. The input signal is a unit amplitude square wave of length $400$ samples and $50\%$ duty cycle. For model estimation, during training phase a Gaussian noise of zero mean and $0.2$ variance was added. }

{The following specifications are used for the RBF algorithms: 
\begin{inparaenum}[(1)]
	\item a learning rate of $1 \times 10^{-4}$,
	\item the Gaussian kernel spread is set to $0.5$, and
	\item for $5$ neurons, the centers are selected as $\textbf{m} = \{-100, -50, 0, 50, -100 \}$.
\end{inparaenum}
}
	\begin{figure}[ht!] 
		\centering
		\includegraphics[width=0.7\textwidth]{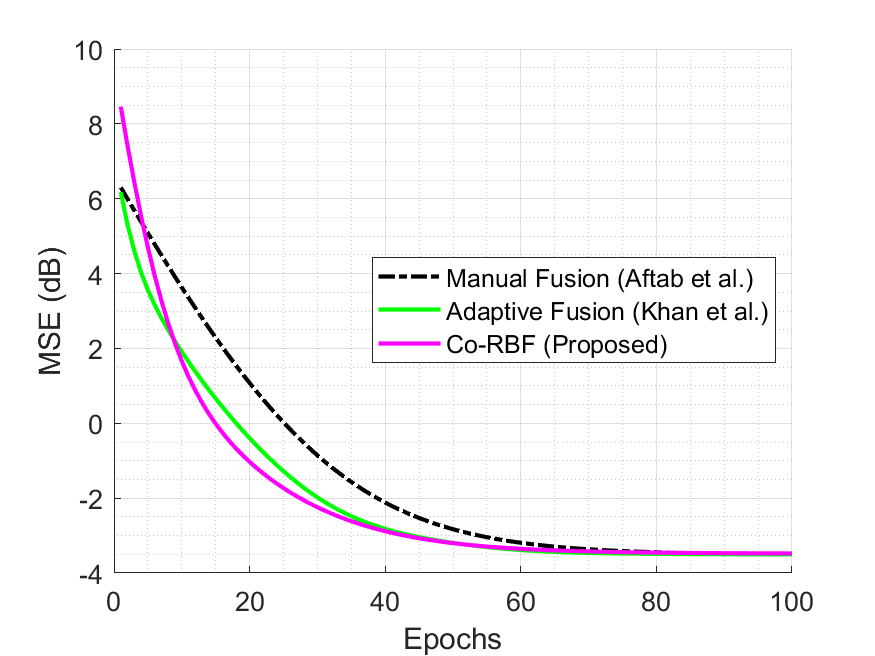}
		\caption{MSE curves of different RBF algorithms on system identification problem. \label{fig:mse_sys_id}} 
	\end{figure}

\begin{figure}[ht!] 
		\centering
		\includegraphics[width=0.7\textwidth]{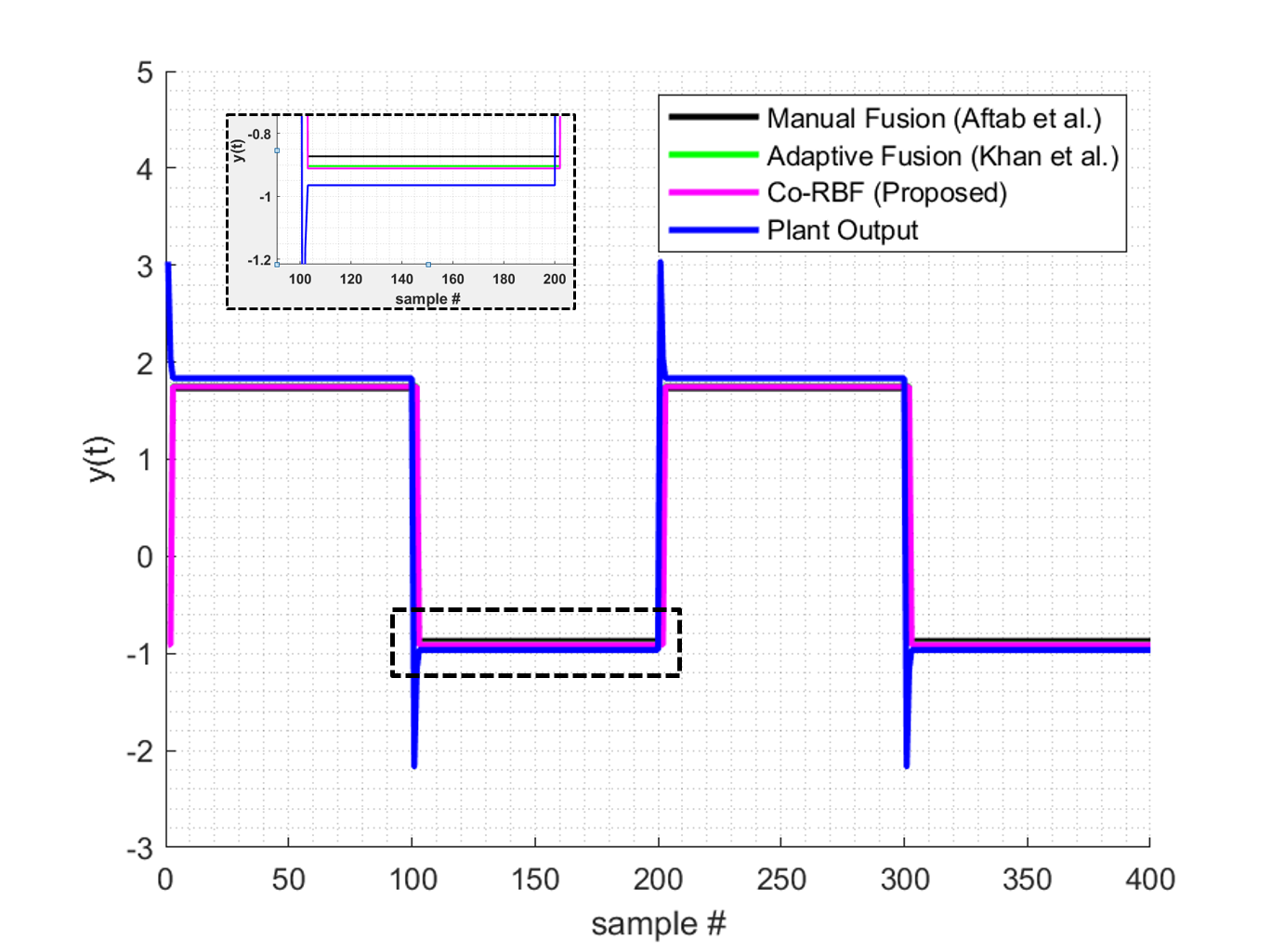}
		\caption{Estimated output of different the RBF algorithms on test data of system identification problem. \label{fig:output_sys_id}} 
	\end{figure}

{MSE curves of different RBF algorithms are depicted in Fig.~\ref{fig:mse_sys_id}. The proposed architecture yields the highest convergence rate with a minimum error of $3.48$ dB which is identical to the manual and adaptive fusion method \cite{aftab_novel_2014,khan2017novel}. Comparison of the actual and estimated test signals for the  different RBF algorithms is illustrated in Fig.~\ref{fig:output_sys_id}. In an inset plot, it is evident that our proposed algorithm estimates the actual test signal significantly better compared to the other algorithms.}

\section{Conclusion}\label{sec:conclusion}
{In this paper, we proposed a novel multi-kernel RBF neural network architecture called Co-RBFNN. The proposed kernel fusion method uses matrix-based mixing weights enabling each (primary and sub-primary) kernel to learn independent weights. A graphical explanation highlighting the underlying reasons for the improvement is provided along with a detailed mathematical analysis. We demonstrated the efficacy of the proposed solution on three important problems, namely: (i) Nonlinear system identification, (ii) pattern classification and (iii) function approximation. The proposed algorithm has shown to comprehensively outperform the two state-of-the-art methods i.e. manual and adaptive fusion of kernels. For the problem of pattern classification, the proposed framework achieved the lowest error floor of $-35.39$ dB after $2000$ epochs of training. For the testing phase the proposed Co-RBFNN achieved a high classification accuracy of $99.13\%$ (approximately) which compares favorably with the contemporary methods. For the function approximation problem, our proposed method converged to the lowest error of $-39.83$ dB after $2000$ epochs. The convergence rate of the proposed algorithm was also found to be better than the competing methods. For the nonlinear system identification problem, the proposed Co-RBFNN algorithm exhibited the fastest convergence rate achieving a minimum error of $-3.48$ dB. The unseen test signal was more accurately estimated by the proposed approach compared to the contemporary methods. MATLAB code for a sample problem can be downloaded from \url{https://github.com/Shujaat123/Robust_RBF}.}
 
 {The proposed novel approach enables independent learning of the mixing weights making it superior compared to the contemporary approaches. However, one sophistication of the current method is that it requires fine-tuning and pre-processing of data, which requires some experience on behalf of inexperienced users. For such users, in future, we are interested in designing a toolbox version that can facilitate the adaptation of the proposed method. Additionally, it would be interesting to incorporate more sophisticated learning strategies such as evolutionary methods and expanding the domain of our experiments to other more practical problems.}

\section*{Acknowledgement}
Syed Muhammad Atif acknowledges the support of HEC, Pakistan under Indigenous Ph.D. Fellowship Program (PIN 315-13358-2EG3-204).

\section*{Author contributions}
All authors contributed to the study conception and design.  Shujaat Khan designed the research, Syed Muhammad Atif and Shujaat Khan conducted and conceived the experiments and performed analysis, The initial draft of the manuscript was written by Syed Muhammad Atif and all authors commented on previous versions of the manuscript. All authors discussed the results and approved the final manuscript.

\bibliographystyle{spmpsci}      
\bibliography{main}

\begin{thebibliography}{10}
\providecommand{\url}[1]{{#1}}
\providecommand{\urlprefix}{URL }
\expandafter\ifx\csname urlstyle\endcsname\relax
  \providecommand{\doi}[1]{DOI~\discretionary{}{}{}#1}\else
  \providecommand{\doi}{DOI~\discretionary{}{}{}\begingroup
  \urlstyle{rm}\Url}\fi

\bibitem{aftab_novel_2014}
Aftab, W., Moinuddin, M., Shaikh, M.S.: A {Novel} {Kernel} for {RBF} {Based}
  {Neural} {Networks}.
\newblock Abstract and Applied Analysis \textbf{2014}, 1--10 (2014).
\newblock \doi{10.1155/2014/176253}.
\newblock \urlprefix\url{http://www.hindawi.com/journals/aaa/2014/176253/}

\bibitem{ahmad2017fclms}
Ahmad, J., Khan, S., Usman, M., Naseem, I., Moinuddin, M., Syed, H.J.: Fclms:
  Fractional complex lms algorithm for complex system identification.
\newblock In: 2017 IEEE 13th International Colloquium on Signal Processing \&
  its Applications (CSPA), pp. 39--43. IEEE (2017)

\bibitem{alexandridis_cooperative_2016}
Alexandridis, A., Chondrodima, E., Sarimveis, H.: Cooperative learning for
  radial basis function networks using particle swarm optimization.
\newblock Applied Soft Computing \textbf{49}, 485--497 (2016).
\newblock \doi{10.1016/j.asoc.2016.08.032}.
\newblock
  \urlprefix\url{https://linkinghub.elsevier.com/retrieve/pii/S1568494616304264}

\bibitem{aljarah_training_2018}
Aljarah, I., Faris, H., Mirjalili, S., Al-Madi, N.: Training radial basis
  function networks using biogeography-based optimizer.
\newblock Neural Computing and Applications \textbf{29}(7), 529--553 (2018).
\newblock \doi{10.1007/s00521-016-2559-2}.
\newblock \urlprefix\url{http://link.springer.com/10.1007/s00521-016-2559-2}

\bibitem{de_almeida_rego_deterministic_2014}
de~Almeida~Rego, J.B., de~Medeiros~Martins, A., Costa, E.d.B.: Deterministic
  {System} {Identification} {Using} {RBF} {Networks}.
\newblock Mathematical Problems in Engineering \textbf{2014}, 1--10 (2014).
\newblock \doi{10.1155/2014/432593}.
\newblock \urlprefix\url{http://www.hindawi.com/journals/mpe/2014/432593/}

\bibitem{bu2020adversarial}
Bu, K., He, Y., Jing, X., Han, J.: Adversarial transfer learning for deep
  learning based automatic modulation classification.
\newblock IEEE Signal Processing Letters  (2020)

\bibitem{bucak_multiple_2014}
Bucak, S.S., Jin, R., Jain, A.K.: Multiple {Kernel} {Learning} for {Visual}
  {Object} {Recognition}: {A} {Review}.
\newblock IEEE Transactions on Pattern Analysis and Machine Intelligence
  \textbf{36}(7), 1354--1369 (2014).
\newblock \doi{10.1109/TPAMI.2013.212}.
\newblock \urlprefix\url{http://ieeexplore.ieee.org/document/6654166/}

\bibitem{chen_combining_2019}
Chen, Z.Y., Kuo, R.J.: Combining {SOM} and evolutionary computation algorithms
  for {RBF} neural network training.
\newblock Journal of Intelligent Manufacturing \textbf{30}(3), 1137--1154
  (2019).
\newblock \doi{10.1007/s10845-017-1313-7}.
\newblock \urlprefix\url{http://link.springer.com/10.1007/s10845-017-1313-7}

\bibitem{fisher_use_1936}
Fisher, R.A.: {THE} {USE} {OF} {MULTIPLE} {MEASUREMENTS} {IN} {TAXONOMIC}
  {PROBLEMS}.
\newblock Annals of Eugenics \textbf{7}(2), 179--188 (1936).
\newblock \doi{10.1111/j.1469-1809.1936.tb02137.x}.
\newblock
  \urlprefix\url{http://doi.wiley.com/10.1111/j.1469-1809.1936.tb02137.x}

\bibitem{fu_sparse_2010}
Fu, L., Zhang, M., Li, H.: Sparse {RBF} {Networks} with {Multi}-kernels.
\newblock Neural Processing Letters \textbf{32}(3), 235--247 (2010).
\newblock \doi{10.1007/s11063-010-9153-x}.
\newblock \urlprefix\url{http://link.springer.com/10.1007/s11063-010-9153-x}

\bibitem{gan_hybrid_2012}
Gan, M., Peng, H., Dong, X.p.: A hybrid algorithm to optimize {RBF} network
  architecture and parameters for nonlinear time series prediction.
\newblock Applied Mathematical Modelling \textbf{36}(7), 2911--2919 (2012).
\newblock \doi{10.1016/j.apm.2011.09.066}.
\newblock
  \urlprefix\url{https://linkinghub.elsevier.com/retrieve/pii/S0307904X11006251}

\bibitem{gao_implementing_2012}
Gao, F., Han, L.: Implementing the {Nelder}-{Mead} simplex algorithm with
  adaptive parameters.
\newblock Computational Optimization and Applications \textbf{51}(1), 259--277
  (2012).
\newblock \doi{10.1007/s10589-010-9329-3}.
\newblock \urlprefix\url{http://link.springer.com/10.1007/s10589-010-9329-3}

\bibitem{goodfellow2016nips}
Goodfellow, I.: Nips 2016 tutorial: Generative adversarial networks.
\newblock arXiv preprint arXiv:1701.00160  (2016)

\bibitem{gu_nonlinear_2016}
Gu, Y., Liu, T., Jia, X., Benediktsson, J.A., Chanussot, J.: Nonlinear
  {Multiple} {Kernel} {Learning} {With} {Multiple}-{Structure}-{Element}
  {Extended} {Morphological} {Profiles} for {Hyperspectral} {Image}
  {Classification}.
\newblock IEEE Transactions on Geoscience and Remote Sensing \textbf{54}(6),
  3235--3247 (2016).
\newblock \doi{10.1109/TGRS.2015.2514161}.
\newblock \urlprefix\url{http://ieeexplore.ieee.org/document/7390058/}

\bibitem{hassan_kernel_2018}
Hassan, A.K., Moinuddin, M., Al-Saggaf, U.M., Shaikh, M.S.: On the {Kernel}
  {Optimization} of {Radial} {Basis} {Function} {Using} {Nelder} {Mead}
  {Simplex}.
\newblock Arabian Journal for Science and Engineering \textbf{43}(6),
  2805--2816 (2018).
\newblock \doi{10.1007/s13369-017-2888-1}.
\newblock \urlprefix\url{http://link.springer.com/10.1007/s13369-017-2888-1}

\bibitem{haykin_neural_1999}
Haykin, S.S.: Neural networks: a comprehensive foundation, 2nd ed edn.
\newblock Prentice Hall, Upper Saddle River, N.J (1999)

\bibitem{haykin_adaptive_2014}
Haykin, S.S.: Adaptive filter theory, fifth edition edn.
\newblock Pearson, Upper Saddle River, New Jersey (2014)

\bibitem{ibrahim2020machine}
Ibrahim, M.S., Dong, W., Yang, Q.: Machine learning driven smart electric power
  systems: Current trends and new perspectives.
\newblock Applied Energy \textbf{272}, 115,237 (2020)

\bibitem{khan2018novel}
Khan, S., Ahmad, J., Naseem, I., Moinuddin, M.: A novel fractional
  gradient-based learning algorithm for recurrent neural networks.
\newblock Circuits, Systems, and Signal Processing \textbf{37}(2), 593--612
  (2018)

\bibitem{khan_spatio-temporal_2018}
Khan, S., Ahmad, J., Sadiq, A., Naseem, I., Moinuddin, M.: Spatio-{Temporal}
  {RBF} {Neural} {Networks}.
\newblock In: 2018 3rd {International} {Conference} on {Emerging} {Trends} in
  {Engineering}, {Sciences} and {Technology} ({ICEEST}), pp. 1--5. IEEE,
  Karachi, Pakistan (2018).
\newblock \doi{10.1109/ICEEST.2018.8643322}.
\newblock \urlprefix\url{https://ieeexplore.ieee.org/document/8643322/}

\bibitem{khan2017flmf}
Khan, S., Ahmed, N., Malik, M.A., Naseem, I., Togneri, R., Bennamoun, M.: Flmf:
  Fractional least mean fourth algorithm for channel estimation in non-gaussian
  environment.
\newblock In: 2017 International Conference on Information and Communication
  Technology Convergence (ICTC), pp. 466--470. IEEE (2017)

\bibitem{khan2019universal}
Khan, S., Huh, J., Ye, J.C.: Universal plane-wave compounding for high quality
  us imaging using deep learning.
\newblock In: 2019 IEEE International Ultrasonics Symposium (IUS), pp.
  2345--2347. IEEE (2019)

\bibitem{khan2020adaptive}
Khan, S., Huh, J., Ye, J.C.: Adaptive and compressive beamforming using deep
  learning for medical ultrasound.
\newblock IEEE Transactions on Ultrasonics, Ferroelectrics, and Frequency
  Control pp. 1--1 (2020)

\bibitem{khan2018fractional}
Khan, S., Naseem, I., Malik, M.A., Togneri, R., Bennamoun, M.: A fractional
  gradient descent-based rbf neural network.
\newblock Circuits, Systems, and Signal Processing \textbf{37}(12), 5311--5332
  (2018)

\bibitem{khan2017novel}
Khan, S., Naseem, I., Togneri, R., Bennamoun, M.: A novel adaptive kernel for
  the rbf neural networks.
\newblock Circuits, Systems, and Signal Processing \textbf{36}(4), 1639--1653
  (2017)

\bibitem{khan2018rafp}
Khan, S., Naseem, I., Togneri, R., Bennamoun, M.: Rafp-pred: Robust prediction
  of antifreeze proteins using localized analysis of n-peptide compositions.
\newblock IEEE/ACM Transactions on Computational Biology and Bioinformatics
  \textbf{15}(1), 244--250 (2018)

\bibitem{lee2019collagan}
Lee, D., Kim, J., Moon, W.J., Ye, J.C.: Collagan: Collaborative gan for missing
  image data imputation.
\newblock In: Proceedings of the IEEE Conference on Computer Vision and Pattern
  Recognition, pp. 2487--2496 (2019)

\bibitem{liu_c-rbfnn:_2018}
Liu, Y., Zhao, J., Xiao, Y.: C-{RBFNN}: {A} user retweet behavior prediction
  method for hotspot topics based on improved {RBF} neural network.
\newblock Neurocomputing \textbf{275}, 733--746 (2018).
\newblock \doi{10.1016/j.neucom.2017.09.015}.
\newblock
  \urlprefix\url{https://linkinghub.elsevier.com/retrieve/pii/S0925231217315102}

\bibitem{meng_nonlinear_2018}
Meng, X., Rozycki, P., Qiao, J.F., Wilamowski, B.M.: Nonlinear {System}
  {Modeling} {Using} {RBF} {Networks} for {Industrial} {Application}.
\newblock IEEE Transactions on Industrial Informatics \textbf{14}(3), 931--940
  (2018).
\newblock \doi{10.1109/TII.2017.2734686}.
\newblock \urlprefix\url{http://ieeexplore.ieee.org/document/7999287/}

\bibitem{muhammad_weighted_2017}
Muhammad, M., Naseem, I., Aftab, W., A~Bencherif, S., Memich, A.: A {Weighted}
  {Cosine} {RBF} {Neural} {Networks}.
\newblock J Mol Biol Biotech \textbf{2}(2), 1--8 (2017).
\newblock
  \urlprefix\url{http://www.imedpub.com/articles/a-weighted-cosine-rbf-neural-networks.pdf}

\bibitem{naseem2017ecmsrc}
Naseem, I., Khan, S., Togneri, R., Bennamoun, M.: Ecmsrc: A sparse learning
  approach for the prediction of extracellular matrix proteins.
\newblock Current Bioinformatics \textbf{12}(4), 361--368 (2017)

\bibitem{nerguizian2006geolocation}
Nerguizian, C., Despins, C., Aff{\`e}s, S.: Geolocation in mines with an
  impulse response fingerprinting technique and neural networks.
\newblock IEEE transactions on wireless communications \textbf{5}(3), 603--611
  (2006)

\bibitem{pal_mountain_2000}
Pal, N.R., Chakraborty, D.: Mountain and subtractive clustering method:
  {Improvements} and generalizations.
\newblock International Journal of Intelligent Systems \textbf{15}(4), 329--341
  (2000).
\newblock \doi{10.1002/(SICI)1098-111X(200004)15:4<329::AID-INT5>3.0.CO;2-9}.
\newblock
  \urlprefix\url{http://doi.wiley.com/10.1002/%28SICI%291098-111X%28200004%2915%3A4%3C329%3A%3AAID-INT5%3E3.0.CO%3B2-9}

\bibitem{peng2018modulation}
Peng, S., Jiang, H., Wang, H., Alwageed, H., Zhou, Y., Sebdani, M.M., Yao,
  Y.D.: Modulation classification based on signal constellation diagrams and
  deep learning.
\newblock IEEE transactions on neural networks and learning systems
  \textbf{30}(3), 718--727 (2018)

\bibitem{pratiwi_mammograms_2015}
Pratiwi, M., {Alexander}, Harefa, J., Nanda, S.: Mammograms {Classification}
  {Using} {Gray}-level {Co}-occurrence {Matrix} and {Radial} {Basis} {Function}
  {Neural} {Network}.
\newblock Procedia Computer Science \textbf{59}, 83--91 (2015).
\newblock \doi{10.1016/j.procs.2015.07.340}.
\newblock
  \urlprefix\url{https://linkinghub.elsevier.com/retrieve/pii/S1877050915018694}

\bibitem{sadiq2018chaotic}
Sadiq, A., Ibrahim, M.S., Usman, M., Zubair, M., Khan, S.: Chaotic time series
  prediction using spatio-temporal rbf neural networks.
\newblock In: 2018 3rd International Conference on Emerging Trends in
  Engineering, Sciences and Technology (ICEEST), pp. 1--5. IEEE (2018)

\bibitem{sadiq2019enhanced}
Sadiq, A., Khan, S., Naseem, I., Togneri, R., Bennamoun, M.: Enhanced q-least
  mean square.
\newblock Circuits, Systems, and Signal Processing \textbf{38}(10), 4817--4839
  (2019)

\bibitem{877615}
{Sikora}, R., {Giza}, Z., {Filipowicz}, F., {Sikora}, J.: The bell function
  approximation of material coefficients distribution in the electrical
  impedance tomography.
\newblock IEEE Transactions on Magnetics \textbf{36}(4), 1023--1026 (2000)

\bibitem{simon_biogeography-based_2008}
Simon, D.: Biogeography-{Based} {Optimization}.
\newblock IEEE Transactions on Evolutionary Computation \textbf{12}(6),
  702--713 (2008).
\newblock \doi{10.1109/TEVC.2008.919004}.
\newblock \urlprefix\url{http://ieeexplore.ieee.org/document/4475427/}

\bibitem{tuia_learning_2010}
Tuia, D., Camps-Valls, G., Matasci, G., Kanevski, M.: Learning {Relevant}
  {Image} {Features} {With} {Multiple}-{Kernel} {Classification}.
\newblock IEEE Transactions on Geoscience and Remote Sensing \textbf{48}(10),
  3780--3791 (2010).
\newblock \doi{10.1109/TGRS.2010.2049496}.
\newblock \urlprefix\url{http://ieeexplore.ieee.org/document/5497137/}

\bibitem{usman2020afp}
Usman, M., Khan, S., Lee, J.A.: Afp-lse: Antifreeze proteins prediction using
  latent space encoding of composition of k-spaced amino acid pairs.
\newblock Scientific Reports \textbf{10}(1), 1--13 (2020)

\bibitem{varma_more_2009}
Varma, M., Babu, B.R.: More generality in efficient multiple kernel learning.
\newblock In: Proceedings of the 26th {Annual} {International} {Conference} on
  {Machine} {Learning} - {ICML} '09, pp. 1--8. ACM Press, Montreal, Quebec,
  Canada (2009).
\newblock \doi{10.1145/1553374.1553510}.
\newblock
  \urlprefix\url{http://portal.acm.org/citation.cfm?doid=1553374.1553510}

\bibitem{vetrivel_disaster_2018}
Vetrivel, A., Gerke, M., Kerle, N., Nex, F., Vosselman, G.: Disaster damage
  detection through synergistic use of deep learning and 3d point cloud
  features derived from very high resolution oblique aerial images, and
  multiple-kernel-learning.
\newblock ISPRS Journal of Photogrammetry and Remote Sensing \textbf{140},
  45--59 (2018).
\newblock \doi{10.1016/j.isprsjprs.2017.03.001}.
\newblock
  \urlprefix\url{https://linkinghub.elsevier.com/retrieve/pii/S0924271616305913}

\bibitem{wen_robust_2019}
Wen, Z., Xie, L., Feng, H., Tan, Y.: Robust fusion algorithm based on {RBF}
  neural network with {TS} fuzzy model and its application to infrared flame
  detection problem.
\newblock Applied Soft Computing \textbf{76}, 251--264 (2019).
\newblock \doi{10.1016/j.asoc.2018.12.019}.
\newblock
  \urlprefix\url{https://linkinghub.elsevier.com/retrieve/pii/S1568494618307087}

\bibitem{yang_fast_2018}
Yang, X., Li, Y., Sun, Y., Long, T., Sarkar, T.K.: Fast and {Robust} {RBF}
  {Neural} {Network} {Based} on {Global} {K}-means {Clustering} with {Adaptive}
  {Selection} {Radius} for {Sound} {Source} {Angle} {Estimation}.
\newblock IEEE Transactions on Antennas and Propagation pp. 1--1 (2018).
\newblock \doi{10.1109/TAP.2018.2823713}.
\newblock \urlprefix\url{http://ieeexplore.ieee.org/document/8335765/}

\bibitem{yang_nature-inspired_2010}
Yang, X.S.: Nature-inspired metaheuristic algorithms, 2. ed edn.
\newblock Luniver Press, Frome (2010)

\bibitem{yoon2018efficient}
Yoon, Y.H., Khan, S., Huh, J., Ye, J.C.: Efficient b-mode ultrasound image
  reconstruction from sub-sampled rf data using deep learning.
\newblock IEEE transactions on medical imaging \textbf{38}(2), 325--336 (2018)

\bibitem{zhu_traffic_2014}
Zhu, J.Z., Cao, J.X., Zhu, Y.: Traffic volume forecasting based on radial basis
  function neural network with the consideration of traffic flows at the
  adjacent intersections.
\newblock Transportation Research Part C: Emerging Technologies \textbf{47},
  139--154 (2014).
\newblock \doi{10.1016/j.trc.2014.06.011}.
\newblock
  \urlprefix\url{https://linkinghub.elsevier.com/retrieve/pii/S0968090X14002010}

\end{thebibliography}

\end{document}